%% file: main.tex
\documentclass[10pt,twocolumn,letterpaper]{article}

\usepackage{cvpr}              %

\input{preamble}

\definecolor{cvprblue}{rgb}{0.21,0.49,0.74}
\usepackage[pagebackref,breaklinks,colorlinks,citecolor=cvprblue]{hyperref}

\usepackage[utf8]{inputenc}
\usepackage[T1]{fontenc}
\usepackage{array}
\usepackage{multirow}
\usepackage{amsmath, bm}
\usepackage{amsfonts}
\usepackage{amssymb}
\usepackage{amsthm}

\usepackage{stmaryrd}
\usepackage{graphicx}
\usepackage[export]{adjustbox}

\usepackage{soul}

\usepackage{bbold}
\usepackage{hyperref}
\usepackage{wrapfig}
\usepackage[table]{xcolor}
\usepackage{enumitem}
\usepackage{changepage,threeparttable}
\usepackage{tabularx}
\usepackage{booktabs}
\usepackage{float}
\hypersetup{colorlinks=true, linkcolor=blue, filecolor=magenta, urlcolor=cyan,}
\usepackage{graphicx}
\usepackage{caption}
\usepackage{comment}

\usepackage{afterpage}
\usepackage{stfloats}

\usepackage{tcolorbox}
\definecolor{lightblue}{rgb}{0.68, 0.85, 0.9}
\definecolor{lightyellow}{rgb}{1.0, 1.0, 0.88}
\definecolor{boxborderblue}{rgb}{0.0, 0.2, 0.6}
\definecolor{boxborderbrown}{rgb}{0.5, 0.25, 0.0}

\theoremstyle{plain}

\theoremstyle{definition}

\theoremstyle{remark}

\setlist[itemize]{topsep=5pt, partopsep=0pt, parsep=2pt, itemsep=5pt}

\def\paperID{345} %
\def\confName{3DV\xspace}
\def\confYear{2026\xspace}

\title{Data-Efficient Inference of Neural Fluid Fields via SciML Foundation Model}

\newcommand*\samethanks[1][\value{footnote}]{\footnotemark[#1]}
\author{
Yuqiu Liu$^1$\thanks{The first two authors contributed equally.} ,\quad%
Jingxuan Xu$^2$\samethanks ,\quad%
Mauricio Soroco$^1$,
Yunchao Wei$^{2,3}$\thanks{Co-corresponding authors.} ,\quad%
Wuyang Chen$^1$\samethanks\\%\hspace{4.5pt}\\
$^1$Simon Fraser University\quad
$^2$Beijing Jiaotong University\quad
$^3$Beijing Academy of Artificial Intelligence
}

\begin{document}
\maketitle

\begin{abstract}
Recent developments in 3D vision have enabled significant progress in inferring \textbf{neural fluid fields} and realistic rendering of fluid dynamics. However, these methods require dense captures of real-world flow, which demand specialized lab setups, making the process costly and challenging.
Scientific machine learning (\textbf{SciML}) \textbf{foundation models}, pretrained on extensive simulations of partial differential equations (PDEs), encode rich multiphysics knowledge and thus provide promising sources of domain priors for inferring fluid fields.
Nevertheless, the transferability of these foundation models to real-world vision problems remains largely underexplored.
In this work, we demonstrate that SciML foundation models can significantly reduce the \textbf{data costs} of inferring real-world 3D fluid dynamics with improved generalization.
Our method leverages strong forecasting capabilities and meaningful representations of SciML foundation models.
We equip neural fluid fields with a novel collaborative training that utilizes augmented frames, and fluid features extracted by our foundation model.
We demonstrate significant advancements in both quantitative metrics and visual quality over previous methods, improving 9$\sim$36\% peak signal-to-noise ratio (PSNR) in future prediction with 25$\sim$50\% reduction in the number of training frames, thereby showcasing the practical applicability of SciML foundation models in real-world fluid dynamics.
We release our code at: \href{https://github.com/delta-lab-ai/SciML-HY}{\ul{https://github.com/delta-lab-ai/SciML-HY}}.
\end{abstract}

\section{Introduction}

\begin{figure}[h!]
\vspace{-1em}
\centering
\includegraphics[width=0.99\linewidth]{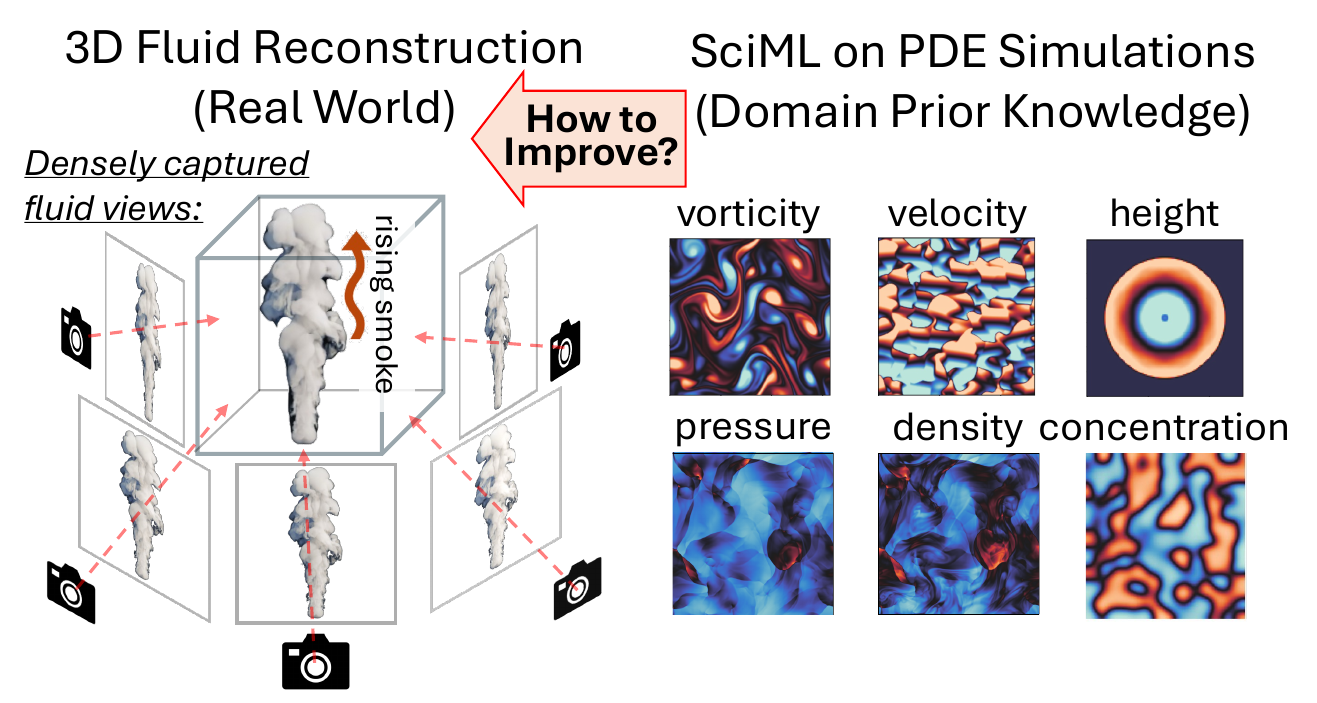}
\centering 
\vspace{-1em}
\caption{
Inferring neural fluid fields requires densely captured views.
Meanwhile, PDE simulations are important for building SciML foundation models.
How to utilize this rich domain knowledge
to improve 3D fluid reconstruction in the real world?
} 
\label{fig:teaser}
\vspace{-1em}
\end{figure}

Fluid phenomena are ubiquitous in our 3D world, from
the powerful ocean currents, to the turbulent jet streams in the air.
One important yet open challenge in understanding fluids is to \ul{recover fluid dynamics from visual observations}, also known as the inference of \textbf{3D fluid fields}.
Formally stated,
given visual inputs (2D images or video sequences),
this task aims to infer invisible quantities like density and velocity
in the spatiotemporal domain (3+1D) (Figure~\ref{fig:teaser} left).
This facilitates downstream rendering of realistic fluids in computer games and videos~\cite{wang2024physics}, and even applications of broad impact, such as
weather forecasting~\cite{pathak2022fourcastnet} and
airfoil design~\cite{thuerey2020deep}.
Unlike rigid bodies, fluids present a unique challenge due to their dynamic and complex nature, requiring advanced computational methods.

Recent advancements in 3D vision have enabled significant progress in inferring fluid fields.
This includes both multi-view benchmarks~\cite{eckert2019scalarflow} of high-quality flow videos with well-calibrated camera poses,
and also neural fluid fields~\cite{chu2022physics,guan2022neurofluid,deng2023fluid,deng2023learning,yu2024inferring} jointly optimized by rendering loss and physics constraints. However, learning neural fluid fields is notorious for its
\textbf{high costs of acquiring dense fluid views}
\footnote{``Frame'' and ``view'' are used interchangeably in the context of fluid field reconstruction, following~\cite{yu2024inferring}. We will unify them under ``frame'' to avoid ambiguity.}.
Methods like HyFluid~\cite{yu2024inferring} require four videos with 120 continuous frames each.
This requirement
relies on \emph{specialized} lab setups.
For example, collecting and calibrating the ScalarFlow dataset~\cite{eckert2019scalarflow} requires insulated containers with heaters, fog machines with servo-controlled valves, and multiple cameras, with an estimated total cost of around \$1,100.
Many fluid dynamics phenomena occur rapidly, necessitating the use of high-speed cameras to capture detailed visualizations. These specialized imaging systems can add significant expenses to experimental setups, with costs reaching thousands of dollars per camera.~\cite{Chronos,zang2020tomofluid}.
With mobile devices or drones, capturing real-world fluid views in the wild will become even more challenging.

A common strategy to achieve data efficiency and improved generalization
is to introduce prior knowledge.
Scientific machine learning (SciML), which aims to learn physical dynamics,
is a promising source of prior (Figure~\ref{fig:teaser}, right).
Deep neural networks (DNNs)
provide surrogate models for approximations of partial differential equations (PDEs) and real-world challenges like weather forecasting~\cite{li2021fourier,pathak2022fourcastnet} and turbulent flow~\cite{kochkov2021machine}.
\textbf{SciML foundation models} are further advanced in recent works~\cite{mccabe2023multiple,hao2024dpot,hang2024unisolver,ye2024pdeformer,wang2024latent,rahman2024pretraining,sun2024towards,shen2024ups}.
By scaling up extensive training datasets to incorporate multiphysical domains and PDE simulations (such as Navier-Stokes, Burgers', shallow water),
SciML foundation models aim to encode common physical behaviors and improve generalization in scientific contexts.

Although promising, SciML foundation models are mainly pretrained and evaluated on \ul{synthetic PDE simulations}~\cite{PDEBench2022,subramanian2023towards,mialon2023self}.
These simulations, while encoding rich physical domain knowledge, still differ from \ul{real fluid captures} with multiscale patterns and noisy measurements (Figure~\ref{fig:teaser}).
This poses questions about the transferability of SciML foundation models in real-world 3D fluid problems.
In contrast, foundation models in popular ML domains have been widely utilized as strong priors.
Vision models such as DINO~\cite{caron2021emerging,oquab2023dinov2} and CLIP~\cite{radford2021learning} have been leveraged to support generalizable representations and semantic awareness~\cite{ye2023featurenerf,tang2023scene,masuda2024generalizable,wang2024lift3d,wang2024d,charatan2024pixelsplat,wewer2024latentsplat}.
Large language models (LLMs)~\cite{achiam2023gpt4,touvron2023llama,touvron2023llama2,dubey2024llama} are pretrained on high-quality corpora and interact with the informal spoken language of human users every day.
Therefore, we ask our core question:

\vspace{-0.5em}
\begin{tcolorbox}[colframe=black, colback=white, boxrule=0.2mm, sharp corners]
\vspace{-0.6em}
\textit{\textbf{How to utilize SciML foundation models to \nobreak advance 3D reconstruction of real-world fluids?}}
\vspace{-0.6em}
\end{tcolorbox}
\vspace{-0.5em}

\begin{figure}[t!]
\centering
\includegraphics[width=0.82\linewidth]{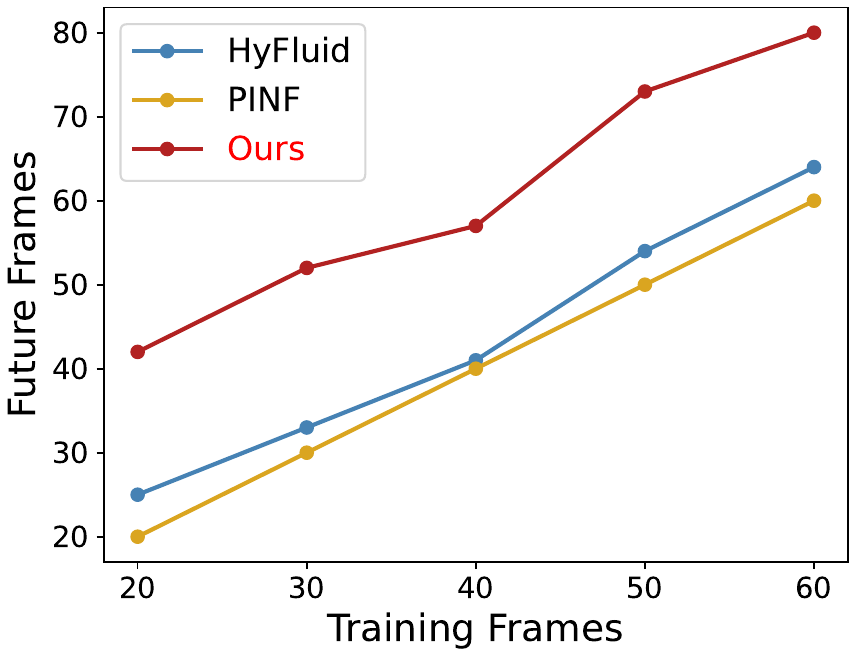}
\centering 
\vspace{-0.5em}
\caption{Our method is significantly more data-efficient than previous works (PINF~\cite{chu2022physics}, HyFluid~\cite{yu2024inferring}) on future prediction.
X-axis: different numbers of training frames (\texttt{nf}) per video.
Y-axis: temporal index of reliably predicted future frames with peak signal-to-noise ratio (PSNR) threshold of 25 (higher is better).}
\label{fig:future_prediction_frames_th25}
\vspace{-1.5em}
\end{figure}

In this work, we provide affirmative answers (Figure~\ref{fig:future_prediction_frames_th25}), and demonstrate that pretrained SciML foundation models can enhance data efficiency in inferring neural fluid fields from sparse videos.
We establish the foundation for incorporating pretrained physics knowledge as a prior for real-world fluid reconstruction. 
Our core idea is to
leverage the strong forecasting and meaningful representations of SciML foundation model, and ``distill'' this prior into neural fluid fields.
We demonstrate both improved quantitative metrics and the high-quality visual appearance of our method on real-world flow captures with significantly reduced training input frames.
Specifically:
\begin{enumerate}[leftmargin=*]
    \item
    Given extremely sparse initial frames from short videos of flows, our foundation model forecasts future steps (temporal frames)
    and enables a collaborative training strategy for neural fluid fields with more augmented fluid frames (Section~\ref{sec:cotraining} and Figure~\ref{fig:collaborative_training}).
    \item
    To improve generalization, we introduce meaningful representations of flows into neural fluid fields.
    These representations are extracted by our foundation model and carefully aligned with the camera rays used in the fluid field (Section~\ref{sec:feature_aggregation} and Figure~\ref{fig:feature_aggregation}).
    \item We provide comprehensive experiments and ablation studies (Section~\ref{sec:exp}). Our method not only unlocks extreme data efficiency (25$\sim$50\% reduction in the number of training frames), but also achieves both improved reconstruction error and visual quality (10$\sim$36\% improved peak signal-to-noise ratio in future prediction).
\end{enumerate}

\section{Preliminary}

\subsection{Inferring Fluid Fields from Videos}

Our method is developed to work with HyFluid~\cite{yu2024inferring}, which infers neural fluid fields (density and velocity) from videos.

\vspace{-1em}
\paragraph{Problem Definition.}
Given videos of smoke rising upwards (Figure~\ref{fig:teaser}, left), with the number of frames (views) used in each video denoted by \texttt{nf}, neural fluid fields aim to infer the density field $\sigma(x, y, z, t)$ and the underlying velocity field $\bm{u}(x, y, z, t) = (v_x, v_y, v_z)$ of the smoke, both parameterized by deep networks.

For the density field $\sigma(x, y, z, t)$, HyFluid randomly samples camera rays $(x, y, z, t)$ and reconstructs the density using a 4D extension of iNGP~\cite{muller2022instant}, which accelerates the neural rendering with multiscale hash encoding of spatiotemporal positions. During training, this density field is optimized by comparing input and rendered views via differentiable volume rendering (Figure~\ref{fig:framework} bottom).
Similarly, the velocity field $\bm{u}(x, y, z, t)$ is inferred by another iNGP model, and is supervised by physics-informed losses that enforce mass conservation for incompressible flows and divergence-free velocity.
We follow the assumptions of the original ScalarFlow dataset, whose reconstruction model assumes incompressible flow (see Section~3 of~\cite{eckert2019scalarflow}). Under atmospheric pressure and low Mach numbers, this is a standard and reasonable approximation for smoke.

During inference, the density field is used to render the visual appearance of the smoke, and the learned velocity field can be used to advect (evolve) the density over temporal steps for both \textit{re-simulation} (interpolation of the temporal range seen during training) and \textit{future prediction} (extrapolation of unseen future temporal ranges).

\vspace{-1em}
\paragraph{The ScalarFlow Dataset: Smoke Videos with Calibrated Cameras.}
Recent works on fluid field reconstruction focus on the ScalarFlow dataset~\cite{eckert2019scalarflow}:
a comprehensive collection of volumetric reconstructions of real-world smoke plumes (Figure~\ref{fig:teaser} left). It encompasses a wide array of complex, buoyancy-driven flows rising upwards that transition into turbulence, capturing observable scalar transport processes.
To the best of our knowledge, the ScalarFlow dataset is by far the best-calibrated benchmark on real-world fluid (smoke) dynamics.

\subsection{SciML Foundation Model}
\label{sec:sciml_fm}

\begin{figure}[t!]
\centering
\includegraphics[width=0.8\linewidth]{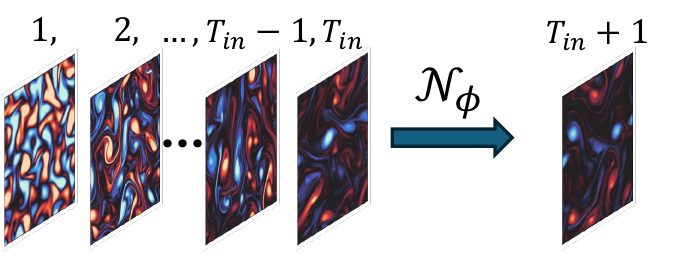}
\centering 
\vspace{-0.5em}
\caption{Forecasting by SciML foundation models~\cite{mccabe2023multiple,hao2024dpot}. Given $T_{in}$ previous steps, the model predicts the next step of the fluid dynamics (here, each frame shows the vorticity of the fluid).} 
\label{fig:fm_forecasting}
\vspace{-1em}
\end{figure}

For time-dependent PDEs, the solution is a mapping from the joint of a spatial and temporal domain to the dynamics of the physical system (e.g. density, velocity, vorticity of the fluid at a certain spatiotemporal location):
$\mathbf{v} := \mathcal{T} \times \mathcal{S} \rightarrow \mathbb{R}^d$.
In current literature~\cite{li2021fourier,li2021physics,PDEBench2022,mccabe2023multiple}, the \textbf{forward modeling} operator $\mathcal{N}$ computes the PDE solution given $T_{in} \in \mathbb{Z}^+$ consecutive previous timesteps:
$\mathcal{N} := \mathbf{v}(t-T_{in} \cdot \Delta t, \cdot), \ldots, \mathbf{v}(t-\Delta t, \cdot) \mapsto \mathbf{v}(t, \cdot)$,
where $\Delta t$ is the granularity of the temporal grid.
This enables finite-difference approximations of the temporal derivatives of PDEs.
See Figure~\ref{fig:fm_forecasting} for an illustration.

SciML aims to find ML-based surrogate models
for forward modeling by learning an approximation from data $\mathcal{N}_{\phi} \simeq \mathcal{N}$ ($\phi$ for learnable parameters).
To optimize $\mathcal{N}_{\phi}$, we take a dataset $\mathcal{D}$ comprising $N$ discretized PDE simulations (``samples'')
$\mathcal{D}:=\left\{\mathbf{v}^{(i)} \mid i=1, \ldots, N\right\}$,
and minimize a loss functional $L$, typically the normalized root of the mean squared error ($\mathrm{nRMSE} \equiv \frac{\left\|\mathbf{v}_{\text {pred }}-\mathbf{v}\right\|_2}{\left\|\mathbf{v} \right\|_2}$ where $\mathbf{v}_{\text {pred }}$ is the prediction from $\mathcal{N}_{\phi}$).

Traditionally, SciML models focus on learning simulations of one PDE~\cite{lu2019deeponet,li2021fourier,li2021physics}.
However, recent works explored and verified benefits of \ul{scaling up the pretraining data to include diverse PDE systems}, thus developing \textbf{SciML foundation models}~\cite{mccabe2023multiple,hao2024dpot,hang2024unisolver,ye2024pdeformer,wang2024latent,rahman2024pretraining,sun2024towards,shen2024ups}.
Intuitively, although these PDEs model very different physical systems, 
this ``multi-tasking'' strategy
1) implicitly enforce the learning of the compositionality of PDEs
(which describe core components like nonlinear advection or diffusion in common and also augment specialized terms like buoyancy or system constraints);
2) facilitate transfer learning and knowledge sharing across multiple PDE families.

\section{Methods}

In our work, we aim to reduce the number of video frames (\texttt{nf}) required by learning neural fluid fields, thereby improving data efficiency. Our method can be applied to any NeRF-based fluid models, in this paper we mainly use HyFluid as our baseline, see the results on other baselines (PINF~\cite{chu2022physics}) in Section~\ref{sec:our_pinf} of the supplementary material.
In Figure~\ref{fig:framework}, we overview our proposed framework\footnote{Our method directly interacts with only the density field of HyFluid, the velocity field is implicitly improved via the density field.}.

\subsection{How to Utilize SciML Foundation Models for Inferring Real-World Fluid Fields?}
\vspace{0.5em}

\begin{figure*}[t!]
\vspace{-0.5em}
\centering
\includegraphics[scale=0.42]{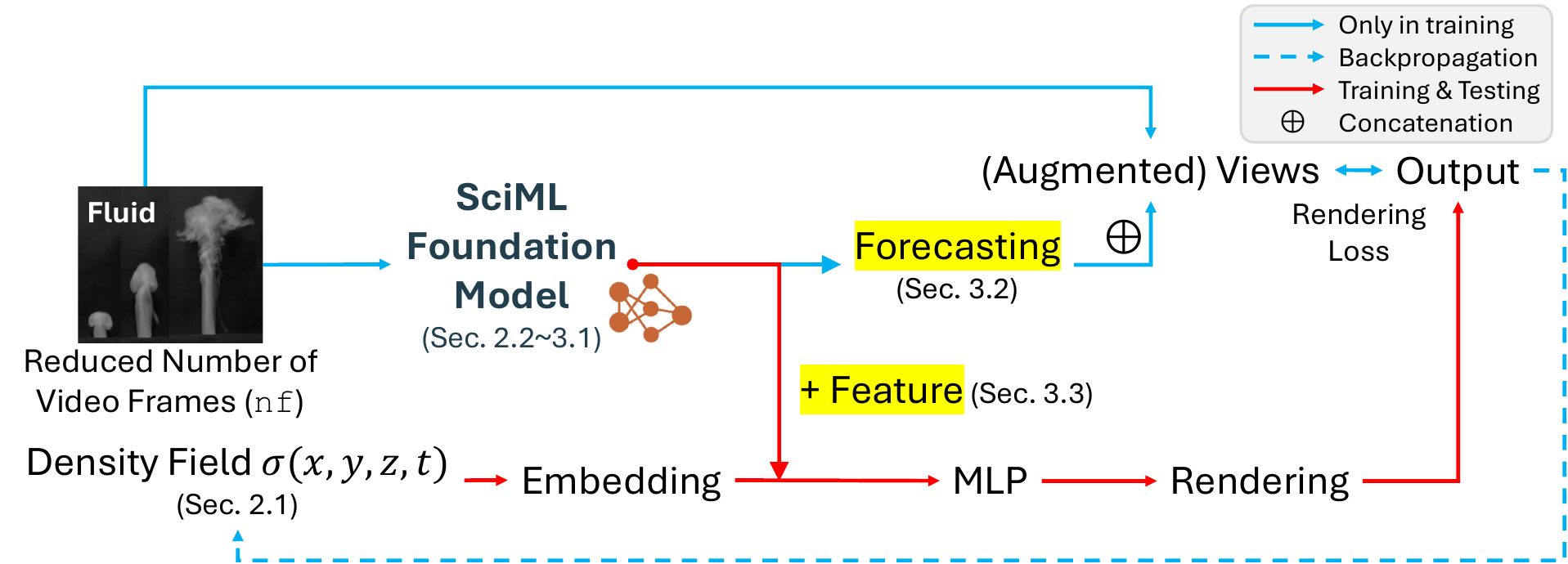}
\centering 
\vspace{-0.5em}
\caption{Overview: We improve the data efficiency (i.e., reduce the number of input fluid frames ``\texttt{nf}'') of learning neural fluid fields via the pretrained SciML foundation model. Given sparse input videos, we utilize our foundation model to: 1) forecast future steps to augment denser frames for training (Section~\ref{sec:cotraining}); 2) extract flow representations and aggregate into embeddings of fluid density fields (Section~\ref{sec:feature_aggregation}).
}
\label{fig:framework}
\vspace{-1.3em}
\end{figure*}

Inspired by recent works~\cite{mccabe2023multiple,hao2024dpot}, we first develop our SciML foundation model as follows:

\begin{enumerate}[leftmargin=*]
\item \textbf{Architecture.}
We adopt a 3D version of the Swin Transformer~\cite{liu2021swin,yang2023swin3d} (6.5M parameters), a popular vision transformer architecture, as our foundation model\footnote{(1) Why not use larger model sizes? We will show that, even with a small model, we can already achieve strong improvements.
Using larger models may further boost
performance,
but this is not the focus of our work.
Recent SciML foundation models also consider sizes smaller than 10M parameters~\cite{wang2024cvit,hao2024dpot,ye2024pdeformer,sun2024towards}.
(2) Architecture choices:
We adhere to the original design of the Swin Transformer and avoid introducing ad hoc modifications. Although recent works on SciML foundation models adopt different architectures~\cite{mccabe2023multiple,hao2024dpot}, the commonly shared aspect of these works is their joint multiphysics pretraining, not their deep network architectures.}. It tokenizes input temporal 2D frames ($\mathbf{v}([t-T_{in}  \cdot \Delta t: t-\Delta t], \cdot)$) with a 3D convolution layer, forwards through efficient windowed attentions, and predicts the next temporal step $\mathbf{v}(t, \cdot)$.
Without loss of generality and following previous works~\cite{PDEBench2022,mccabe2023multiple,hao2024dpot}, we choose $T_{in}=10$. Tuning $T_{in}$ may lead to better performance, but is not the focus of our method.

\item \textbf{Multiphysics Pretraining.}
\phantomsection\label{multiphysics_pretrainin}
We utilize the PDEBench dataset~\cite{PDEBench2022} for pretraining.
Specifically, we pretrain our foundation model on the \ul{joint of diverse simulations} of the following PDEs: both compressible and incompressible Navier-Stokes, shallow water, and reaction-diffusion.
See Section \ref{sec:sciml} in the supplement for details.%
We sample each equation uniformly, zero-pad channels of PDEs with fewer variables, and match different PDE simulations to the same spatial resolution via interpolation.
We will experimentally verify the benefits of this multiphysics pretraining in Section~\ref{sec:exp_benefits_pretraining}.

\item \textbf{Fine-tuning.}
After pretraining, we fine-tune on ScalarFlow\footnote{The same set of sparse video frames that will be used to train HyFluid.}.
Inspired by recent works~\cite{lam2022graphcast,pathak2022fourcastnet}, we employ a curriculum schedule to encourage forecasting further temporal steps, gradually increasing autoregressive steps from 3 to 8 by 1 every 20 training epochs.
Both pretraining and fine-tuning use the $\mathrm{nRMSE}$ loss.

\end{enumerate}

\vspace{-0.2em}
We expect \textbf{two core benefits} of our SciML foundation model that can be utilized in the real world (highlighted with yellow in Figure~\ref{fig:framework}):

\begin{enumerate}[leftmargin=*]
\item \textbf{Strong Forecasting.}
As our foundation model is pretrained with the next-frame prediction, it can natively forecast precise future steps as augmented frames of fluids to complement sparse videos (Section~\ref{sec:cotraining}).

\item \textbf{Representation Learning.}
As a data-driven approach similar to DINO~\cite{caron2021emerging,oquab2023dinov2}, the feature space constructed by our SciML foundation model can extract meaningful features of fluids to facilitate better generalization of 3D neural fluid fields (Section~\ref{sec:feature_aggregation}).
\end{enumerate}

\subsection{Co-Training via Foundation Model Forecasting}
\label{sec:cotraining}

\begin{figure}[h!]
\vspace{-1em}
\centering
\includegraphics[width=0.92\linewidth]{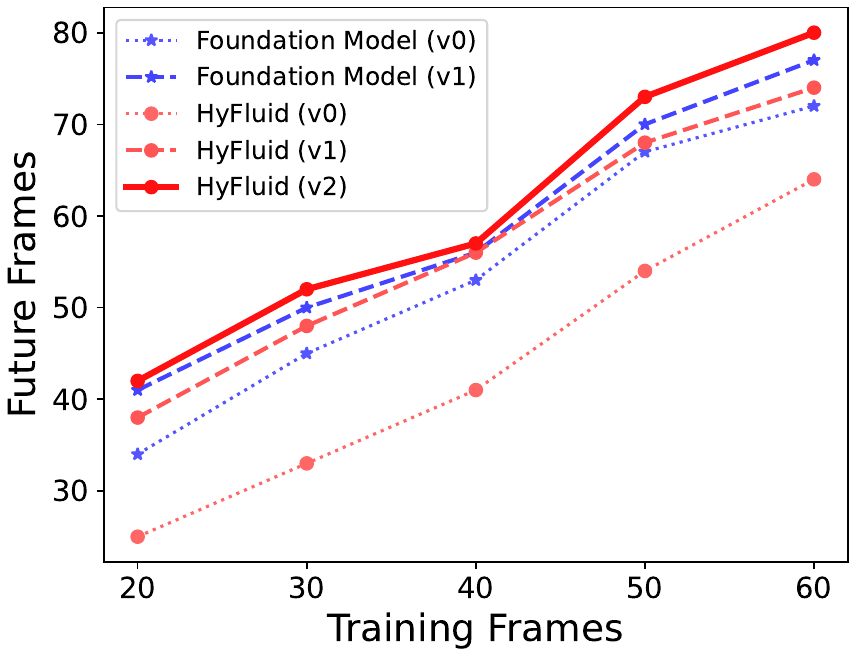}
\centering 
\vspace{-1em}
\caption{Collaborative training between HyFluid and the foundation model improves future predictions.
``v0, v1, v2'' match models annotated in Figure~\ref{fig:collaborative_training}.
HyFluid can be progressively improved (v0$\rightarrow$v1$\rightarrow$v2) with more augmented frames.
Y-axis: temporal index of reliably predicted future frames (thresholded by PSNR=25).
X-axis: number of training frames (\texttt{nf}) per video.
}
\label{fig:fm_hyfluid_cotraining}
\vspace{-1em}
\end{figure}

Given sparse smoke videos, one way to address data scarcity is to augment more frames.
We first study the forecasting performance of both our SciML foundation model and the neural fluid fields.
As shown by two dotted curves in Figure~\ref{fig:fm_hyfluid_cotraining} (``Foundation Model v0'' vs. ``HyFluid v0''), the forecasting quality of the foundation model is much better than the neural fluid fields.

\begin{figure}[b!]
\vspace{-2em}
\centering
\includegraphics[scale=0.5]{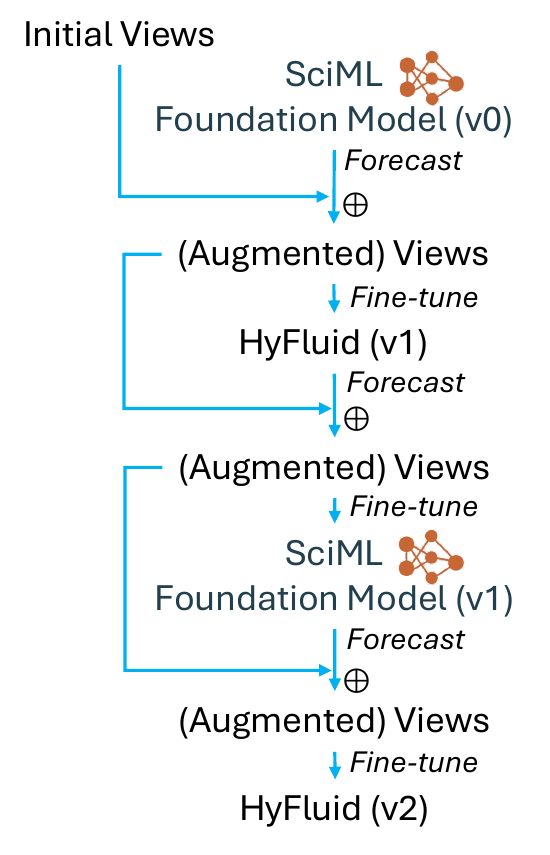}
\centering
\vspace{-1em}
\caption{Collaborative training between HyFluid and our SciML foundation model via forecasting with augmented frames. ``v0, v1, v2'' match the corresponding curves in Figure~\ref{fig:fm_hyfluid_cotraining}.}
\label{fig:collaborative_training}
\vspace{-1em}
\end{figure}

To utilize the strong forecasting of our foundation model, we propose a collaborative training strategy for neural fluid fields.
The core idea is to train the foundation model and neural fluid field with augmented frames (Figure~\ref{fig:collaborative_training}).
We alternately concatenate the reliably predicted frames (thresholded by PSNR=25) from the foundation model or neural fluid field into the current training set, and fine-tune each other.
This collaborative training can also be viewed as a knowledge distillation from the foundation model to the neural fluid field in the output space.
As shown by two dashed curves in Figure~\ref{fig:fm_hyfluid_cotraining} (``Foundation Model v1'', ``HyFluid v1''), the collaborative training enhances the forecasting of both models, and the final version of the neural fluid field (solid curve ``HyFluid v2'') achieves much stronger future predictions.
By achieving comparable PSNR with fewer input frames, we demonstrate that our collaborative training can significantly improve the data efficiency of neural fluid fields.
Notably, while the foundation model itself is data-driven and does not explicitly encode the Navier–Stokes equation, the training of HyFluid (v2) over extended temporal frames still introduces new physical knowledge through regularization from fluid simulation.

\subsection{Feature Aggregation from Foundation Model}
\label{sec:feature_aggregation}

\begin{figure}[b!]
\vspace{-1.5em}
\centering
\includegraphics[scale=0.46]{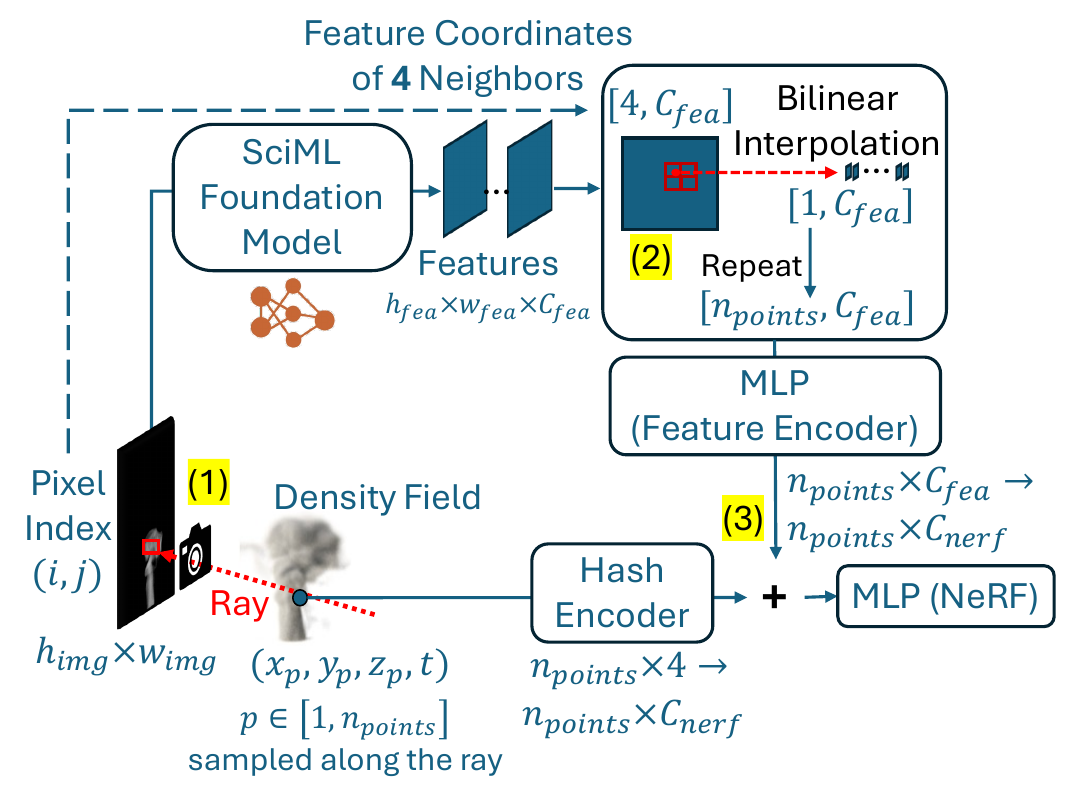}
\centering 
\vspace{-1.5em}
\caption{We aggregate representations learned by our SciML foundation model into HyFluid with three steps: (1) project from spatiotemporal location to the camera plane; (2) extract and interpolate neighboring features; (3) aggregate features from the foundation model into neural fields.
``$C_{fea}$'': feature dimension of SciML foundation model.
``$C_{nerf}$'': hidden dimension of neural density field (NeRF).
}
\label{fig:feature_aggregation}
\vspace{-1em}
\end{figure}

In addition to leveraging the augmented frames via the foundation model's forecasting, we further aggregate the learned representation from the foundation model into the neural fluid field.
This can be viewed as a knowledge distillation from the foundation model to the neural fluid field in the feature space.
We show our design of feature aggregation in Figure~\ref{fig:feature_aggregation}.
This includes three steps:
\begin{enumerate}[leftmargin=*]
\item For each camera ray $(x_p, y_p, z_p)$, we use the camera's extrinsics and intrinsics to project the ray onto the position in image coordinates $(h_{img}, w_{img})$.
\item We reshape the sequence of tokens in our foundation model into 2D feature maps, and extract the feature vector corresponding to the camera ray via interpolating over the neighboring four feature coordinates.
This feature vector is shared by all points sampled along the ray.
\item We use a two-layer MLP (with ReLU activation) to map the feature vector to the same feature dimensionality as the embedded features of the spatiotemporal coordinates of the density field, and sum them for aggregation.
\end{enumerate}

During training, features are extracted from fluid frames from videos.
During testing, since videos are not accessible, the SciML foundation model extracts features based on frames rendered by the density field from prior temporal steps.
To extract features of frames before the temporal step at $T_{in}$, we use temporal-wise interpolation to supplement necessary frames as inputs to the foundation model.

\section{Experiments}
\label{sec:exp}

\subsection{Settings}
\label{sec:exp_settings}
\paragraph{Datasets.}
We use real captures from the ScalarFlow dataset~\cite{eckert2019scalarflow}, released in the repository of HyFluid.
For each scene, there are five videos from five cameras fixed at positions evenly distributed across a 120$^\circ$ arc centered at the rising smoke.
In each video, we consider the first \texttt{nf} frames (where the smoke plumes upwards from the bottom), and adjust \texttt{nf} to study our data efficiency.
Each video has a resolution of $1920 \times 1080$.
These videos have been post-processed to remove backgrounds.
Following HyFluid, for each scene, we use four videos for training and hold one out for testing (i.e., as the ground-truth novel view).

\begin{table*}[t!]
\centering
\vspace{-1em}
\caption{Comparing PSNR (higher the better) of fluid field reconstruction by different methods.
We report mean values over 3 random runs (see Table~\ref{table:psnr_nframes_std} in the supplement for standard deviations).
``\texttt{nf}'': number of input training frames (views).
For future prediction, we report the PSNR averaged over 20 future frames (i.e., frames with indices from \texttt{nf}+1 $\rightarrow$ \texttt{nf}+20).}

\vspace{-1em}
\resizebox{1.0\textwidth}{!}{
\addtolength{\tabcolsep}{-0.5em}
\begin{tabular}{lccccccccc}
\toprule
\multirow{2}{*}{Methods} & \multicolumn{3}{c}{Novel View Synthesis} & \multicolumn{3}{c}{Re-Simulation} & \multicolumn{3}{c}{Future Prediction} \\ \cmidrule{2-4} \cmidrule{5-7} \cmidrule{8-10}
 & \texttt{nf}=20 & \texttt{nf}=40 & \texttt{nf}=60 & \texttt{nf}=20 & \texttt{nf}=40 & \texttt{nf}=60 & \texttt{nf}=20 & \texttt{nf}=40 & \texttt{nf}=60 \\ \midrule

PINF~\cite{chu2022physics}    & 33.45  & 31.05  & 30.90  & 24.28  & 24.86  & 24.08  & 21.71  & 20.85  & 20.67  \\  
HyFluid~\cite{yu2024inferring} & 33.83 (+0.38) & 33.32 (+2.27) & \textbf{32.84} (+1.94)  & 33.89 (+9.61) & 33.27 (+8.41) & 32.02 (+7.94)  & 25.22 (+3.51) & 23.98 (+3.13)  & 23.66 (+2.99) \\  
Ours            & \textbf{34.50} (+1.05) & \textbf{33.48} (+2.43) &  \textbf{32.84} (+1.94) & \textbf{34.34} (+10.06)   & \textbf{33.36} (+8.50)  &  \textbf{32.42} (+8.34) & \textbf{27.59} (+5.88)  & \textbf{28.36} (+7.51)  & \textbf{27.76} (+7.09) \\ \bottomrule

\end{tabular}
}
\label{table:psnr_nframes}
\end{table*}

\begin{figure*}[t!]
\vspace{-1.2em}
\centering
\includegraphics[width=0.85\linewidth]{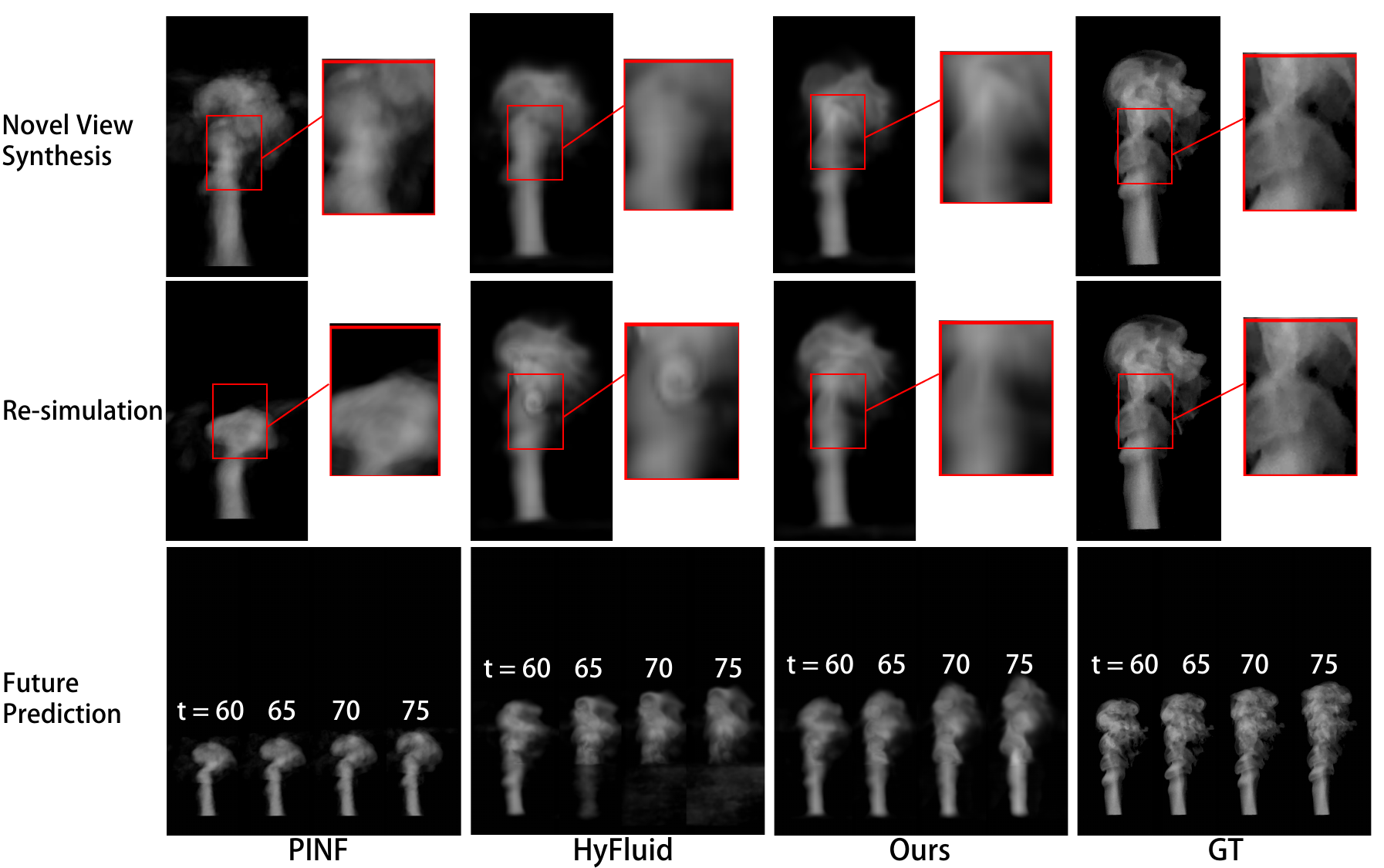}
\centering 
\vspace{-0.5em}
\caption{Visualization of novel view synthesis (top), re-simulation (middle), future prediction (bottom) on ScalarFlow~\cite{eckert2019scalarflow} when 60 frames (per video) are used for training (i.e., \texttt{nf}=60). ``GT'': ground truth.
} 
\vspace{-1.5em}
\label{fig:visualizations}
\end{figure*}

\vspace{-1.2em}
\paragraph{Tasks.}
We compare with two previous works on neural fluid fields: PINF~\cite{chu2022physics} and HyFluid~\cite{yu2024inferring}.
Due to the lack of true 3D volume in ScalarFlow, we evaluate the reconstruction quality using view rendering.
Following~\cite{yu2024inferring}, we consider three tasks:
novel view synthesis, re-simulation, and future prediction.
In novel view synthesis, the density field is used to render smoke views from unseen camera parameters; thus, its quality is evaluated based on rendering accuracy.
For re-simulation and future prediction, the learned velocity field is utilized to advect the density across temporal coordinates. Thus, the quality of the learned velocity field is assessed based on its effect on the density field.
In our future prediction experiments, no model is ever trained with ground-truth future frames from videos.
We refer the reader to~\cite{yu2024inferring} for more details about these tasks.

\vspace{-1.2em}
\paragraph{Evaluation Metrics.}
We report the peak signal-noise ratio (PSNR) averaged over frames.
We leave the structural similarity index measure (SSIM) and the perceptual metric LPIPS~\cite{zhang2018unreasonable} in Section \ref{sec:ssim} in the supplementary material.
These metrics are also widely used in previous deblurring works~\cite{nah2017deep,kupyn2018deblurgan}.

\subsection{Data-Efficiency Inference of Fluid Fields}
\label{sec:exp_quantity}

We first report the inference of fluid density fields.
By default, HyFluid~\cite{yu2024inferring} and PINF~\cite{chu2022physics} used 120 frames (i.e., \texttt{nf}=120) from each video during training.
We consider using a much fewer numbers of sparse training frames than HyFluid and PINF.
During collaborative training, we use 20 augmented frames (from \texttt{nf}+1 to \texttt{nf}+20) in each round. These predicted frames are refreshed rather than accumulated. 
Our method improves both data efficiency and performance. 
As shown in Table~\ref{table:psnr_nframes}, our PSNR consistently outperforms HyFluid and PINF under different sparse training frames (\texttt{nf}), across all three tasks\footnote{PSNRs across different \texttt{nf}s are not comparable, since the numbers of testing frames used to calculate PSNR are also adjusted to be equal to the numbers of training frames.}.
Most importantly, in future prediction,
our method can improve PSNR by
9\% (27.59 vs. 25.22)
and up to 36\% (28.36 vs. 20.85)
compared to HyFluid and PINF, respectively.
This strong and reliable future prediction further contributes to a 25$\sim$50\% reduction in the number of training frames (Figure~\ref{fig:future_prediction_frames_th25}), achieving significant data efficiency.

\subsection{More Realistic Visual Quality} 
\label{sec:exp_quality}

Besides measuring PSNR, it is crucial to visually assess the rendering quality of different methods to ensure realistic and artifact-free reconstructions.
We present qualitative comparisons in Figure~\ref{fig:visualizations}.
For both novel view synthesis and re-simulation, our method successfully recovers fine-grained details while mitigating artifacts in HyFluid and PINF.
Our ability to accurately reconstruct density fields further unlocks high-fidelity future predictions.
Even when provided with sparse input frames, our method is significantly more stable and robust than HyFluid and PINF, which suffer from degraded reconstructions and weak forecasting capabilities.
Our approach preserves the original structure of the fluid while maintaining a natural and physically consistent upward flow.
Both the PSNR measurements and qualitative visualizations strongly indicate that our reconstruction is quantitatively superior and visually more realistic than those produced by HyFluid and PINF.

\begin{table*}[t!]
\centering
\caption{Benefit of multiphysics pretraining on the PSNR (higher the better) of fluid field reconstruction.
``\texttt{nf}'': number of input training frames.%
For future prediction, we report the PSNR averaged over 20 future frames (i.e., frames with indices from \texttt{nf}+1 $\rightarrow$ \texttt{nf}+20).}
\vspace{-1em}
\resizebox{0.80\textwidth}{!}{
\begin{tabular}{lccccccccc}
\toprule
\multirow{2}{*}{Methods} & \multicolumn{3}{c}{Novel View Synthesis} & \multicolumn{3}{c}{Re-Simulation} & \multicolumn{3}{c}{Future Prediction} \\ \cmidrule{2-4} \cmidrule{5-7} \cmidrule{8-10}
 & \texttt{nf}=20 & \texttt{nf}=40 & \texttt{nf}=60 & \texttt{nf}=20 & \texttt{nf}=40 & \texttt{nf}=60 & \texttt{nf}=20 & \texttt{nf}=40 & \texttt{nf}=60 \\ \midrule
No Pretraining & 34.77  & 32.83 & 32.29 &34.12 & 32.97 & \textbf{32.51} & 26.58  & 25.92  & 26.61 \\

+Multiphysics Pretraining & \textbf{34.50} & \textbf{33.48} & \textbf{32.84} & \textbf{34.34} & \textbf{33.36} & 32.42 & \textbf{27.59} & \textbf{28.36} & \textbf{27.76} \\ \bottomrule  

\end{tabular}
}
\label{table:psnr_nframes_pretrain}
\vspace{-0.5em}
\end{table*}

\subsection{Benefits of SciML Pretraining}
\label{sec:exp_benefits_pretraining}

As the core of our SciML foundation model is the joint pretraining on diverse PDE simulations (Section~\ref{multiphysics_pretrainin}),
it is critical to study and verify the true benefits of the domain knowledge from multiphysics pretraining.

To compare different PDE sources, we also pretrain another SciML model on the Maxwell equations, which govern electromagnetic waves and largely differ from the Navier-Stokes equations of ScalarFlow\footnote{Simulation settings for Maxwell are in Section~\ref{pdes}.}.

We analyze and identify two benefits below.

\begin{figure}[t!]
\vspace{-0.5em}
\centering
\includegraphics[width=0.82\linewidth]{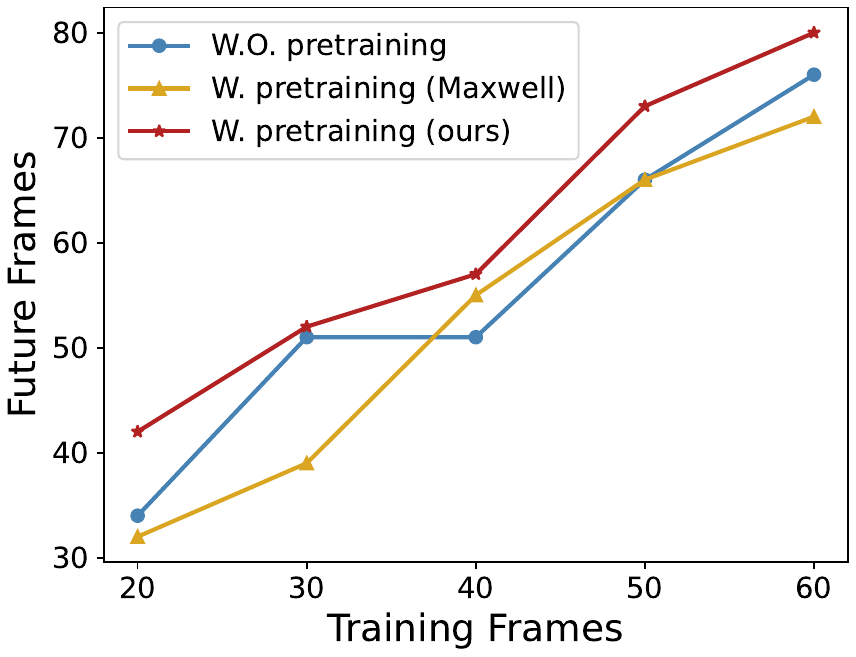}
\centering 
\vspace{-1em}
\caption{Benefit of multiphysics pretraining on future prediction over different numbers of initial training frames per input video (x-axis). We show the temporal index of reliably predicted future frames (thresholded by PSNR=25) on the y-axis (higher is better).
}
\label{fig:future_prediction_frames_th25_pretrain}
\vspace{-1.5em}
\end{figure}

\vspace{-1.2em}
\paragraph{Improved generalization of neural fluid fields.}

We compare the performance of neural fluid fields equipped with our foundation model, with and without multiphysics pretraining.
As shown in Figure~\ref{fig:future_prediction_frames_th25_pretrain}, our multiphysics pretraining can largely improve the data efficiency of neural fluid fields during future prediction.
In contrast, both SciML models—without pretraining or pretrained on irrelevant PDE simulations (Maxwell)—lead to worse future predictions. 
Moreover, over all three fluid reconstruction tasks, the utilization of multiphysics pretraining leads to much improved PSNR, as shown in Table~\ref{table:psnr_nframes_pretrain}. We also evaluate DPOT~\cite{hao2024dpot} with pretrained weights, see results in Section ~\ref{sec:dpot} in the supplementary material. 
These results validate the necessity of high-quality pretraining of our SciML foundation model, and the lack of a strong prior is the key to the worse performance of HyFluid and PINF.

\begin{figure}[t]
\vspace{-1em}
\centering
\includegraphics[width=0.82\linewidth]{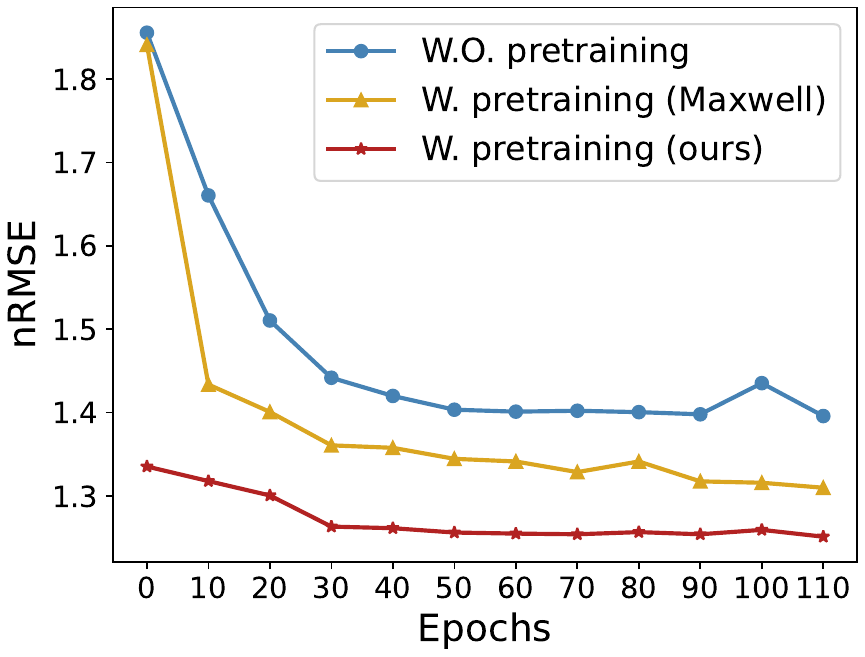}
\centering 
\vspace{-1em}
\caption{Multiphysics pretraining accelerates the convergence during fine-tuning of our SciML foundation model (on 40 initial frames from each of the four training videos in ScalarFlow).} 
\label{fig:convergence_finetuning}
\vspace{-0.5em}
\end{figure}

\vspace{-1.2em}
\paragraph{Faster convergence during fine-tuning.}
Multiphysics pretraining also enables fast convergence during fine-tuning on real-world fluid data.
As shown in Figure~\ref{fig:convergence_finetuning}, despite gaps between PDE simulations and Scalarflow, our pretrained weights can still be quickly adapted to achieve accurate predictions and forecasting.
In comparison, SciML models without pretraining or pretrained on Maxwell converge much more slowly during fine-tuning.

\begin{figure}[t!]
\vspace{-0.5em}
\centering
\includegraphics[width=0.82\linewidth]{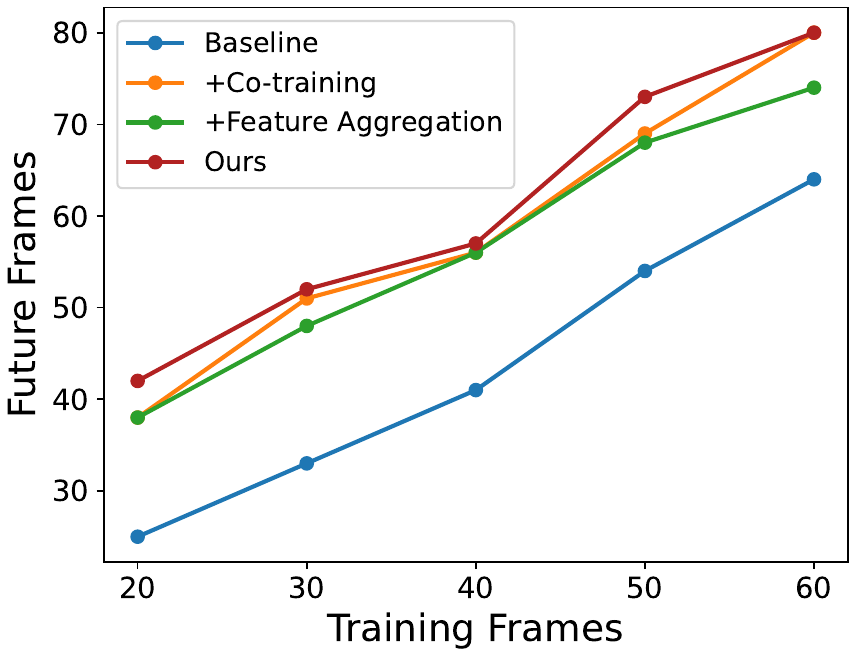}
\centering 
\vspace{-1em}
\caption{Ablation study of our decomposed methods on future prediction.
X-axis: different numbers of initial training frames per video.
Y-axis: temporal index of reliably predicted future frames (thresholded by PSNR=25) (higher is better).}
\label{fig:future_prediction_frames_th25_ablation}
\vspace{-1.5em}
\end{figure}

\subsection{Ablation Study}
\label{sec:exp_ablation}

\begin{table*}[t!]
\centering
\caption{Ablation study of our methods on the PSNR (higher the better) of fluid field reconstruction.
``\texttt{nf}'': number of input training frames.
For future prediction, we report the PSNR averaged over 20 future frames (i.e., frames with indices from \texttt{nf}+1 $\rightarrow$ \texttt{nf}+20).}
\vspace{-1em}
\resizebox{0.84\textwidth}{!}{
\begin{tabular}{lccccccccc}
\toprule
\multirow{2}{*}{Methods} & \multicolumn{3}{c}{Novel View Synthesis} & \multicolumn{3}{c}{Re-Simulation} & \multicolumn{3}{c}{Future Prediction} \\ \cmidrule{2-4} \cmidrule{5-7} \cmidrule{8-10}
 & \texttt{nf}=20 & \texttt{nf}=40 & \texttt{nf}=60 & \texttt{nf}=20 & \texttt{nf}=40 & \texttt{nf}=60 & \texttt{nf}=20 & \texttt{nf}=40 & \texttt{nf}=60 \\ \midrule
Baseline (HyFluid~\cite{yu2024inferring})  & 33.52  & 32.12 & 31.64 & 33.27 & 32.98  & 31.56 & 23.91 & 23.98  & 23.84 \\
+ Co-training & \textbf{34.56} & 33.19 & 32.31 & 34.03 & 33.13 & \textbf{32.61} & \textbf{28.02} & 25.20 & 26.87 \\
+ Feature Aggregation & 33.88 & 33.18 & 32.76 & 33.88 & 33.29 & 32.09 & 26.58 & 27.13 & 25.61 \\  
Ours & 34.50 & \textbf{33.48} & \textbf{32.84} & \textbf{34.34} & \textbf{33.36} & 32.42 & 27.59 & \textbf{28.36} & \textbf{27.76} \\ \bottomrule  

\end{tabular}
}
\label{table:psnr_nframes_ablation}
\vspace{-1.2em}
\end{table*}

We further provide ablation studies on our decoupled framework to demonstrate the benefits of each individual component. As shown in Table~\ref{table:psnr_nframes_ablation} and Figure~\ref{fig:future_prediction_frames_th25_ablation}, both our methods outperform the HyFluid baseline, with the combined approach achieving the best performance.
For more ablation studies and comparison with other SciML foundation models, please read Section~\ref{sec:more_results} in the supplementary material.

\section{Related works}

\subsection{3D Reconstruction of Fluid}

To reconstruct 3D fluid from visual measurements,
traditional approaches utilized active sensing~\cite{hawkins2005acquisition,gu2012compressive,ji2013reconstructing}
or particle imaging velocimetry (PIV)
~\cite{elsinga2006tomographic,adrian2011particle}.
While effective, they necessitated sophisticated and controlled lab environments.
Supervised view synthesis was recently proposed.
In~\cite{zang2020tomofluid}, regularizers on view interpolation and projection consistency were designed for reconstruction from light tomography views.
NeRFlow~\cite{du2021neural} learned 4D spatiotemporal representations of dynamic scenes by capturing 3D occupancy, radiance, and dynamics while enforcing consistency across different modalities.
PINF~\cite{chu2022physics} reconstructed fluid dynamics by leveraging PDEs (Navier-Stokes) to train a continuous spatiotemporal scene representation with a neural radiance field. 
NeuroFluid~\cite{guan2022neurofluid} proposed a particle-driven neural renderer that integrates fluid physical properties into volume rendering and includes a particle transition model to minimize differences between rendered and observed fluid views.
HyFluid~\cite{yu2024inferring} and FluidNexus~\cite{gao2025fluidnexus} jointly learned fluid density and velocity fields, using a set of physics-based losses to enforce physically plausible density and velocity fields.
WillSmoke~\cite{liu2025wildsmoke} further extended smoke reconstruction to real-world scenarios.
However, no existing works explored the introduction of prior knowledge for data efficiency and improve generalization.

\vspace{-1em}
\subsection{Scientific Machine Learning}

SciML, fueled by advancements in deep learning, models physical phenomena and differential equations~\cite{lagaris1998artificial,lagaris2000neural,chen1995universal,chen1995approximation}.
Physics-informed neural networks (PINNs)~\cite{raissi2019physics,zhu2019physics,geneva2020modeling,gao2021phygeonet,ren2022phycrnet} aim to incorporate physics into neural networks by including the differential form of the PDE as a physics-informed regularization term. 
However, this paradigm has been confined to specific PDE scenarios (e.g., fixed PDE coefficients).
Moreover, recent work has highlighted several fundamental issues with PINN-based methods\cite{krishnapriyan2021characterizing,EdwCACM22}.
In contrast, operator learning methods, including Fourier Neural Operators~\cite{li2021fourier,li2020multipole,kovachki2023neural} and the Deep Operator Network~\cite{lu2019deeponet}, have made progress in approximating the solution operators of PDEs.
Although these data-driven approaches show promise in learning PDE solutions, they rely on vast quantities of high-fidelity labeled data.
Researchers have also explored generating synthetic PDE solutions to train SciML models~\cite{hasani2024generating,chen2024data,xu2025hybrid,ma2026learning}.
More recently, SciML foundation models have been developed~\cite{mccabe2023multiple,hao2024dpot,hang2024unisolver,ye2024pdeformer,wang2024latent,rahman2024pretraining,sun2024towards,shen2024ups} by scaling up training datasets to incorporate multiple PDE simulations.
SciML foundation models aim to encode common physical behaviors and enhance the generalization and scalability of SciML.

\subsection{Foundation Models for 3D Reconstruction}

Foundation models for vision are large-scale models pretrained on vast amounts of images or videos, designed to generalize across downstream vision tasks
~\cite{caron2021emerging,radford2021learning,ramesh2021zero,xie2021self,zhi2021place,ramesh2022hierarchical,rombach2022high,wei2022masked}.
CLIP~\cite{radford2021learning} employed contrastive learning with extensive image-text data and achieved zero-shot performance.
DINO~\cite{caron2021emerging} exemplified self-supervised learning and achieved impressive segmentation with minimal supervision, proving useful for visual correspondence and recognition~\cite{amir2021deep,choudhury2021unsupervised,melas2022deep,wang2022self}.
Diffusion models~\cite{ho2020denoising,dhariwal2021diffusion,nichol2021improved,rombach2022high} demonstrated exceptional image generation capabilities, with their learned feature spaces also serving recognition purposes, such as in semantic segmentation~\cite{baranchuk2021label,wolleb2022diffusion}.
To leverage these 2D vision foundation models in 3D reconstruction,
researchers increasingly explored the potential of distilling 2D features into 3D space, exemplified by generalizable neural radiance fields (NeRFs) proposed to bridge this gap~\cite{ye2023featurenerf,tang2023scene,wang2024d,charatan2024pixelsplat,wewer2024latentsplat,masuda2024generalizable,wang2024lift3d,zhang2024condense}.
Fine-tuning pretrained source models while training 3D reconstruction is also a common strategy. For example, Condense~\cite{zhang2025condense} enhances downstream task performance by jointly pretraining 2D and 3D features through multi-view images, creating a unified 2D-3D feature embedding space.
From a broader perspective, when applying DINO to downstream domains with significantly different data distributions, such as medical imaging~\cite{cui2024surgical} or image matting~\cite{yao2024vitmatte}, further fine-tuning is often necessary.

\section{Conclusions}

In this work, we demonstrate that integrating SciML foundation models with neural fluid fields provides a substantial improvement in data efficiency and generalization for inferring 3D fluid fields. Through a collaborative training approach, our method leverages the foundation model's forecasting capabilities to augment data, thereby reducing the reliance on extensive training frames. Additionally, the aggregation of pretrained representations enables more accurate reconstructions of fluid dynamics from sparse video frames. The results indicate that this strategy not only enhances reconstruction quality but also achieves robust performance in novel view synthesis and future prediction. Our work highlights the practical applicability of SciML foundation models in real-world fluid dynamics.

{
    \small
    \bibliographystyle{ieeenat_fullname}
    \bibliography{main}
}

\appendix
\clearpage
\setcounter{page}{1}
\maketitlesupplementary

\input{X_suppl}

\end{document}

%% file: preamble.tex
\usepackage[dvipsnames]{xcolor}

%% file: X_suppl.tex
\section{More Results}
\label{sec:more_results}

\subsection{Justifications for Using PSNR=25 as the Threshold for Reliable Future Predictions}

\begin{figure}[h!]
\centering
\centering
\includegraphics[width=0.48\linewidth]{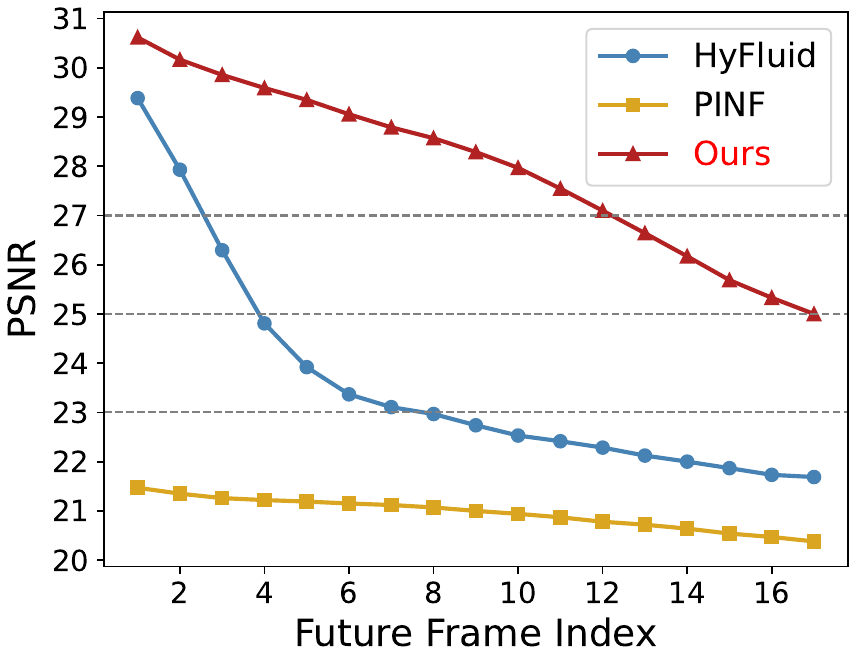}
\includegraphics[width=0.48\linewidth]{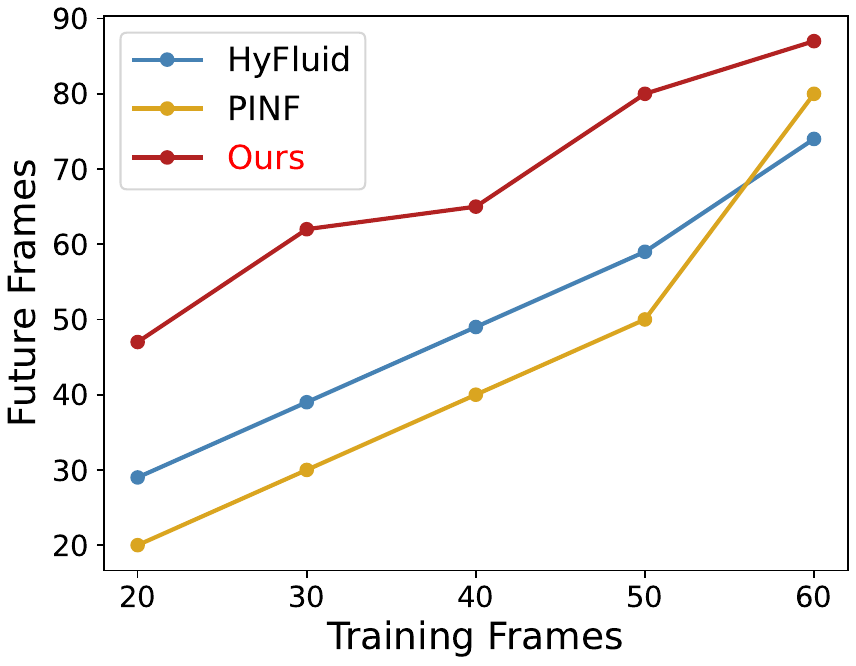}
\includegraphics[width=0.48\linewidth]{images/nf_compare_pinf_psnr25.pdf}
\includegraphics[width=0.48\linewidth]{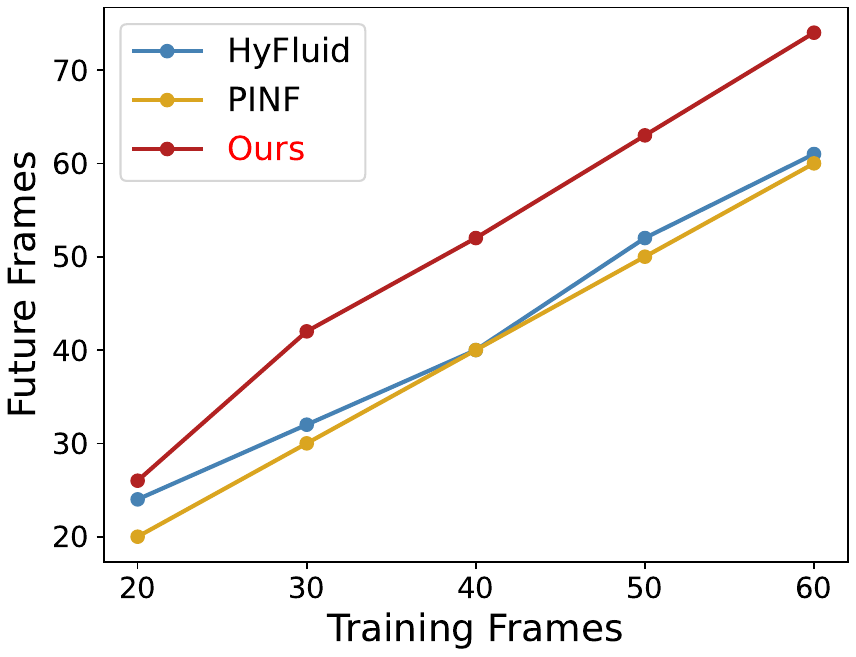}
\centering 
\caption{Justification for using PSNR=25 as the threshold for reliable future predictions. \ul{Top left}: PSNRs of predicted future frames naturally decrease over longer time steps (results obtained with initial training frames \texttt{nf}=40). \ul{Other three plots}: number of predicted future frames (y-axis) thresholded with different PSNR, \ul{23 (top right), 25 (bottom left), 27 (bottom right)}, over different numbers of initial training frames per input video (x-axis).} 
\label{fig:future_prediction_psnrs}
\end{figure}

\begin{figure*}[b!]
\centering
\includegraphics[width=0.75\linewidth]{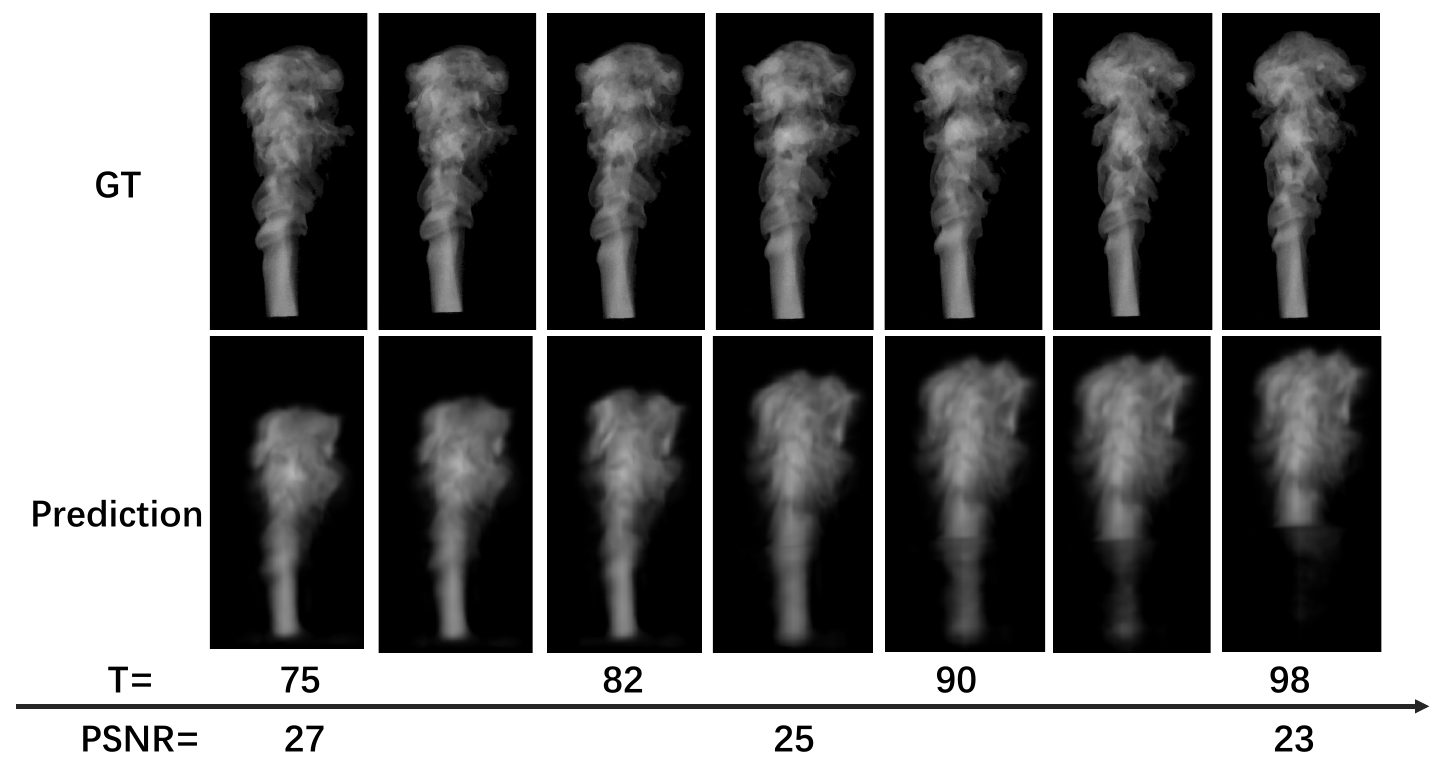}
\centering 
\vspace{-0.5em}
\caption{Justification for using PSNR=25 as the threshold by comparing visualizations of our predicted future frames with different PSNRs. From left to right: PSNR progressively drops from 27 to 23. We use 60 initial frames (per video) for training in this example (i.e., \texttt{nf}=60). ``GT'': ground truth.
} 
\label{fig:visualize_future_prediction_psnr_23_25_27}
\end{figure*}

In our work, we mainly use PSNR=25 as the threshold for determining reliable future predictions.
As shown in Figure~\ref{fig:future_prediction_psnrs} top left, the PSNR of forecasted frames decreases when we make further predictions, suggesting different possible PSNR thresholds (i.e., any future frames of PSNR below the threshold will be considered as unreliable with bad quality).
Here, we provide justifications that this threshold is appropriate on ScalarFlow.

First, as shown in three subplots to the right in Figure~\ref{fig:future_prediction_psnrs}, our data efficiency and improvements persist with different PSNR thresholds, indicating that using different thresholds will not change our conclusion.

Second, when we visualize the rendering quality of future frames with different PSNRs (Figure~\ref{fig:visualize_future_prediction_psnr_23_25_27}), we can see that, both frames with PSNR 25 and 27 are of high visual quality, whereas the frame with PSNR=23 is much visually worse.
This implies that choosing the threshold as PSNR = 25 is sufficient to characterize reliable reconstructions, but 23 already indicates bad qualities.

\subsection{HyFluid with More Network Parameters}

To demonstrate that our improvements are from the pretraind SciML foundation model instead of extra network parameters, we further design and compare with a HyFluid of a larger model size.
Our SciML foundation model has 6.5M parameters.
The original density or velocity field in HyFluid~\cite{yu2024inferring} each has 14.9M parameters.
We train an enlarged HyFluid model with both the density and velocity fields increased to 20.5M parameters.

\begin{figure}[h!]
\vspace{-0.5em}
\centering
\includegraphics[width=0.75\linewidth]{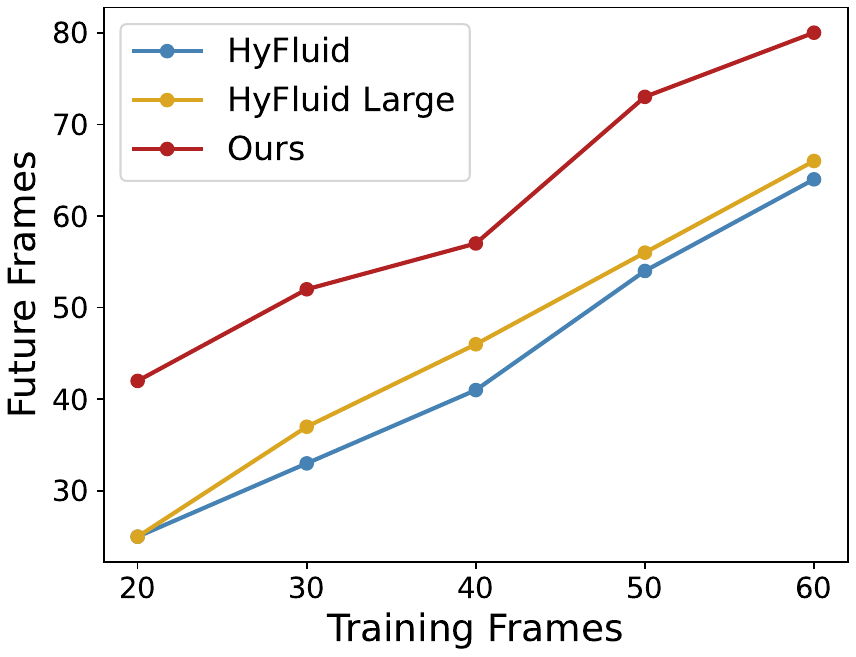}
\centering 
\vspace{-0.5em}
\caption{Comparing with \ul{HyFluid of even a larger model size}, our method is still more data-efficient on future prediction, over different numbers of initial training frames per video (x-axis). We show the temporal index of reliably predicted future frames, using a PSNR threshold of 25, on the y-axis.}
\label{fig:nf_compare_pinf_psnr25_larger_hyfluid}
\vspace{-0.5em}
\end{figure}

As shown in Table~\ref{table:psnr_nframes_larger_hyfluid} and Figure~\ref{fig:nf_compare_pinf_psnr25_larger_hyfluid}, our method can largely outperform HyFluid with a naively enlarged model size, confirming that our benefit is not rooted in more network parameters. The MLP used to aggregate features from the foundation model into the neural field contains only approximately 8K parameters, contributing a negligible increase in model size.

\begin{table*}[t!]
\centering
\caption{PSNR (higher the better) of \ul{HyFluid with more network parameters}.
``\texttt{nf}'': number of input training frames.
For future prediction, we report the PSNR averaged over 20 future frames (i.e., frames with indices from \texttt{nf}+1 $\rightarrow$ \texttt{nf}+20).}
\resizebox{1.0\textwidth}{!}{
\addtolength{\tabcolsep}{-0.35em}
\begin{tabular}{l cccc ccccccccc}
\toprule
\multirow{2}{*}{Methods} 

& \multirow{2}{*}{\begin{tabular}{@{}c@{}}\#Params.\\(M)         \end{tabular}} 
& \multirow{2}{*}{\begin{tabular}{@{}c@{}}Inference\\ Memory (G)   \end{tabular} }
& \multirow{2}{*}{\begin{tabular}{@{}c@{}}Inference\\Speed (FPS)     \end{tabular} }
& \multirow{2}{*}{\begin{tabular}{@{}c@{}}Training Time\\(GPU Hour)    \end{tabular}}

& \multicolumn{3}{c}{Novel View Synthesis} & \multicolumn{3}{c}{Re-Simulation} & \multicolumn{3}{c}{Future Prediction} \\ \cmidrule{6-8} \cmidrule{9-11} \cmidrule{12-14}

& & & & & \texttt{nf}=20 & \texttt{nf}=40 & \texttt{nf}=60 & \texttt{nf}=20 & \texttt{nf}=40 & \texttt{nf}=60 & \texttt{nf}=20 & \texttt{nf}=40 & \texttt{nf}=60 \\ \midrule

HyFluid~\cite{yu2024inferring}  & 29.8 & ~33 & 2.22 & ~17 & 33.52  & 32.12 & 31.64 & 33.27 & 32.98  & 31.56 & 23.91 & 23.98  & 23.84 \\

HyFluid Large & 41 (+37.6\%)  & ~36 & 2.05 & ~20  &34.47  & 32.37 & 32.84 & 34.16 &  33.02 & \textbf{32.47} &  24.10 & 24.21 & 24.06 \\
Ours       & 36.3 (+21.8\%) & ~34 & 2.1 & ~20       &  \textbf{34.50} & \textbf{33.48} & \textbf{32.84} & \textbf{34.34} & \textbf{33.36} & 32.42 & \textbf{27.59} & \textbf{28.36} & \textbf{27.76} \\ \bottomrule  
\end{tabular}
}
\label{table:psnr_nframes_larger_hyfluid}
\end{table*}

\subsection{Random Seeds and Standard Deviations}

We further compare different methods considering standard deviations from random runs, in Table~\ref{table:psnr_nframes_std}.
Our confidence is comparable with previous methods.

\begin{table*}[t!]
\centering
\caption{We report \ul{standard deviations} (3 random runs) of PSNR (higher the better) by different methods.
``\texttt{nf}'': number of input training frames.
For future prediction, we report the PSNR averaged over 20 future frames (i.e., frames with indices from \texttt{nf}+1 $\rightarrow$ \texttt{nf}+20).}
\vspace{-0.5em}
\resizebox{1.0\textwidth}{!}{
\addtolength{\tabcolsep}{-0.35em}
\begin{tabular}{lccccccccc}
\toprule
\multirow{2}{*}{Methods} & \multicolumn{3}{c}{Novel View Synthesis} & \multicolumn{3}{c}{Re-Simulation} & \multicolumn{3}{c}{Future Prediction} \\ \cmidrule{2-4} \cmidrule{5-7} \cmidrule{8-10}
 & \texttt{nf}=20 & \texttt{nf}=40 & \texttt{nf}=60 & \texttt{nf}=20 & \texttt{nf}=40 & \texttt{nf}=60 & \texttt{nf}=20 & \texttt{nf}=40 & \texttt{nf}=60 \\ \midrule
PINF~\cite{chu2022physics}    & 33.45 ($\pm$ 0.22) & 31.05 ($\pm$ 1.14) &  30.90 ($\pm$ 0.44) &24.28 ($\pm$ 0.00) & 24.86 ($\pm$ 0.06) & 24.08 ($\pm$ 0.07) &  21.71 ($\pm$ 0.00) & 20.85 ($\pm$ 0.01) &  20.67 ($\pm$ 0.03) \\
HyFluid~\cite{yu2024inferring} &  33.83 ($\pm$ 0.08) & 33.32 ($\pm$ 0.06) & 32.84 ($\pm$ 0.14)  & 33.89 ($\pm$ 0.62) & 33.27 ($\pm$ 0.23) & 32.02 ($\pm$ 0.20)  &  25.22 ($\pm$ 1.04)& 23.98 ($\pm$ 0.05)  & 23.66 ($\pm$ 0.12) \\
Ours            & \textbf{34.50} ($\pm$ 0.44) & \textbf{33.48} ($\pm$ 0.20) &  \textbf{32.84} ($\pm$ 0.14) & \textbf{34.34} ($\pm$ 0.65)   & \textbf{33.36} ($\pm$ 0.07)  &  \textbf{32.42} ($\pm$ 0.23) & \textbf{27.59} ($\pm$ 0.24)  & \textbf{28.36} ($\pm$ 0.87)  & \textbf{27.76} ($\pm$ 0.09) \\ \bottomrule
\end{tabular}
}
\vspace{-0.5em}
\label{table:psnr_nframes_std}
\end{table*}

\begin{table*}[t!]
\centering
\caption{The PSNR (higher the better) of our method outperforms previous works \ul{over more initial numbers of training frames} (\texttt{nf}).
For future prediction, we report the PSNR averaged over 20 future frames (i.e., frames with indices from \texttt{nf}+1 $\rightarrow$ \texttt{nf}+20).}
\resizebox{1.0\textwidth}{!}{
\begin{tabular}{lccccccccc}
\toprule
\multirow{2}{*}{Methods} & \multicolumn{3}{c}{Novel View Synthesis} & \multicolumn{3}{c}{Re-Simulation} & \multicolumn{3}{c}{Future Prediction} \\ \cmidrule{2-4} \cmidrule{5-7} \cmidrule{8-10}
 & \texttt{nf}=70 & \texttt{nf}=80 & \texttt{nf}=90 & \texttt{nf}=70 & \texttt{nf}=80 & \texttt{nf}=90 & \texttt{nf}=70 & \texttt{nf}=80 & \texttt{nf}=90 \\ \midrule

PINF~\cite{chu2022physics}    & 31.20 & 30.18 & 29.73 & 24.15 & 24.32 & 25.02 & 21.09 & 21.75 & 22.90 \\
HyFluid~\cite{yu2024inferring} & 32.58 ($+1.38$) & 32.28 ($+2.10$) & 32.39 ($+2.66$) & 31.82 ($+7.67$) & 31.48 ($+7.16$) & 30.47 ($+5.45$) & 21.77 ($+0.68$) & 20.00 ($-1.75$) & 21.28 ($-1.62$) \\
Ours            & \textbf{32.88} ($+1.68$) & \textbf{32.55} ($+2.37$) & \textbf{32.47} ($+2.74$) & \textbf{31.99} ($+7.84$) & \textbf{31.60} ($+7.28$) & \textbf{31.14} ($+6.12$) & \textbf{30.94} ($+9.85$) & \textbf{26.62} ($+4.87$) & \textbf{26.62} ($+3.72$) \\ \bottomrule
\end{tabular}
}
\label{table:psnr_nframes_more}
\end{table*}

\begin{table*}[t]
\vspace{-0.5em}
\centering
\caption{Comparing average PSNR (higher the better) of fluid field reconstruction by different methods on other ScalarFlow videos.
``\texttt{nf}'': number of input training frames (views).
For future prediction, we report the PSNR averaged over 20 future frames (i.e., frames with indices from \texttt{nf}+1 $\rightarrow$ \texttt{nf}+20).}
\vspace{-0.5em}
\resizebox{0.7\textwidth}{!}{
\begin{tabular}{lccccccccc}
\toprule
\multirow{2}{*}{Methods} & \multicolumn{3}{c}{Novel View Synthesis} & \multicolumn{3}{c}{Re-Simulation} & \multicolumn{3}{c}{Future Prediction} \\ \cmidrule{2-4} \cmidrule{5-7} \cmidrule{8-10}
 & \texttt{nf}=20 & \texttt{nf}=40 & \texttt{nf}=60 & \texttt{nf}=20 & \texttt{nf}=40 & \texttt{nf}=60 & \texttt{nf}=20 & \texttt{nf}=40 & \texttt{nf}=60 \\ \midrule
HyFluid~\cite{yu2024inferring} & 36.01 & 36.16 & 35.71 & 37.09 &  35.54 &  34.81 & 26.53 & 26.37 & 26.25 \\
Ours            & \textbf{37.49} & \textbf{36.63} & \textbf{36.06} & \textbf{37.50} & \textbf{35.88} & \textbf{35.15} & \textbf{28.12} & \textbf{30.12} & \textbf{28.93} \\ \bottomrule
\end{tabular}
}
\vspace{-0.5em}
\label{table:psnr_nframes_all_video}
\end{table*}

\subsection{Results on Other ScalarFlow Videos}
Beyond the set of video released in HyFluid~\cite{yu2024inferring},
we further provide averaged PSNR of our method and HyFluid on other 2 of ScalarFlow videos, in Table~\ref{table:psnr_nframes_all_video}. Our method also demonstrates stable superiority on other ScalarFlow videos. %

\subsection{Training with More Initial frames}

\begin{figure}[h!]
\centering
\includegraphics[width=0.8\linewidth]{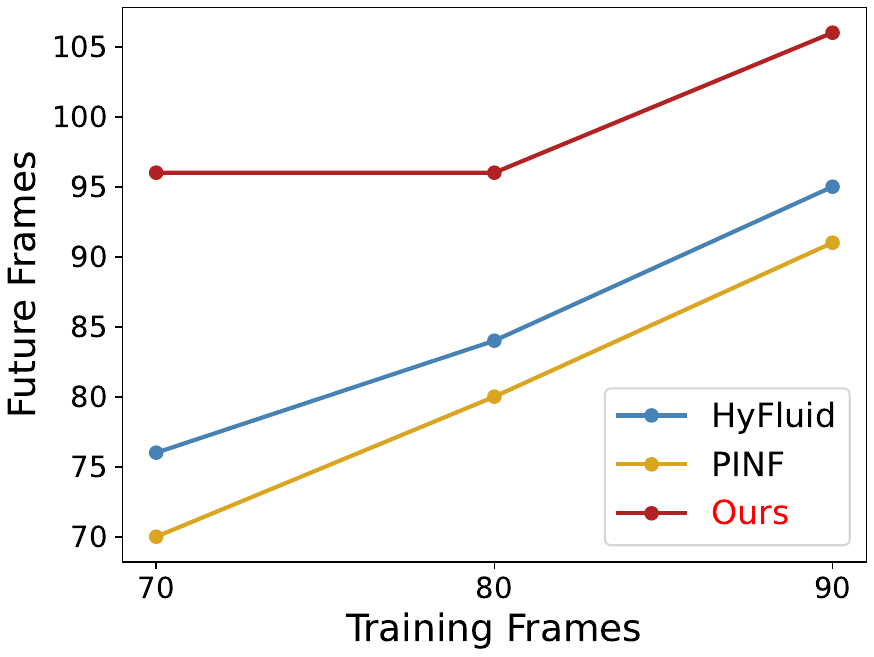}
\centering 
\vspace{-0.5em}
\caption{With \ul{more initial training frames per video} (\texttt{nf}, x-axis), our method is still consistently more data-efficient compared with previous works (PINF~\cite{chu2022physics}, HyFluid~\cite{yu2024inferring}) on future prediction. We show the temporal index of reliably predicted future frames, using a PSNR threshold of 25, on the y-axis.}
\label{fig:nf_compare_pinf_psnr25_more_nf}
\vspace{-0.5em}
\end{figure}

We further consider more numbers of initial frames (larger \texttt{nf}).
We show our results in Table~\ref{table:psnr_nframes_more} and Figure~\ref{fig:nf_compare_pinf_psnr25_more_nf}.
Again, our method consistently outperforms HyFluid~\cite{yu2024inferring} and PINF~\cite{chu2022physics} on more initial frames.

\subsection{SSIM and LPIPS}
\label{sec:ssim}
Beyond PSNR, we further report the structural similarity index measure (SSIM) in Table~\ref{table:ssim_nframes} and the perceptual metric LPIPS~\cite{zhang2018unreasonable} in Table~\ref{table:lpips_nframes}.
Our method is still shown to perform better based on these two additional metrics.

\begin{table*}[t!]
\centering
\caption{Comparing \ul{SSIM} (higher the better) of fluid field reconstruction by different methods.
``\texttt{nf}'': number of input training frames.
For future prediction, we report the SSIM averaged over 20 future frames (i.e., frames with indices from \texttt{nf}+1 $\rightarrow$ \texttt{nf}+20).}
\resizebox{0.75\textwidth}{!}{
\begin{tabular}{lccccccccc}
\toprule
\multirow{2}{*}{Methods} & \multicolumn{3}{c}{Novel View Synthesis} & \multicolumn{3}{c}{Re-Simulation} & \multicolumn{3}{c}{Future Prediction} \\ \cmidrule{2-4} \cmidrule{5-7} \cmidrule{8-10}
 & \texttt{nf}=20 & \texttt{nf}=40 & \texttt{nf}=60 & \texttt{nf}=20 & \texttt{nf}=40 & \texttt{nf}=60 & \texttt{nf}=20 & \texttt{nf}=40 & \texttt{nf}=60 \\ \midrule
PINF~\cite{chu2022physics} &   0.9688  & 0.9648 & 0.9412 &  0.9759 & 0.9530 & 0.9346	 &  0.9471 &  \textbf{0.9046} &\textbf{0.8796} \\
HyFluid~\cite{yu2024inferring}  &  0.9770 & \textbf{0.9741} & 0.9616 &  0.9678 & 0.9667 & 0.9665  &0.9086  & 0.8872  & 0.8686\\
Ours    & \textbf{0.9812} & 0.9644 & \textbf{0.9645} & \textbf{0.9818} & \textbf{0.9677}  &  \textbf{0.9693} & \textbf{0.9526}& 0.8963  &0.8752 \\ \bottomrule
\end{tabular}
}
\label{table:ssim_nframes}
\end{table*}

\begin{table*}[t!]
\centering
\caption{Comparing \ul{LPIPS}~\cite{zhang2018unreasonable} (smaller the better) of fluid field reconstruction by different methods.
``\texttt{nf}'': number of input training frames.
For future prediction, we report the LPIPS averaged over 20 future frames (i.e., frames with indices from \texttt{nf}+1 $\rightarrow$ \texttt{nf}+20).}
\resizebox{0.75\textwidth}{!}{
\begin{tabular}{lccccccccc}
\toprule
\multirow{2}{*}{Methods} & \multicolumn{3}{c}{Novel View Synthesis} & \multicolumn{3}{c}{Re-Simulation} & \multicolumn{3}{c}{Future Prediction} \\ \cmidrule{2-4} \cmidrule{5-7} \cmidrule{8-10}
 & \texttt{nf}=20 & \texttt{nf}=40 & \texttt{nf}=60 & \texttt{nf}=20 & \texttt{nf}=40 & \texttt{nf}=60 & \texttt{nf}=20 & \texttt{nf}=40 & \texttt{nf}=60 \\ \midrule
PINF~\cite{chu2022physics} &  0.0454&0.0501 & 0.0682 &0.0605   &0.0728  & 0.0889	 & 0.1066 & 0.1396  & 0.1539   \\
HyFluid~\cite{yu2024inferring} & 0.0323 & \textbf{0.0395} & 0.0611 & 0.0385 & \textbf{0.0470} & 0.0637  & 0.0922 &   0.1228 & 0.1489 \\
Ours   &  \textbf{0.0293} & 0.0478 & \textbf{0.0567} &  \textbf{0.0296}&  0.0474 & \textbf{0.0565 } &\textbf{0.0625} &\textbf{0.1075} &\textbf{0.1315} \\ \bottomrule
\end{tabular}
}
\label{table:lpips_nframes}
\end{table*}

\subsection{Other SciML Foundation Model}

\label{sec:dpot}
\begin{figure}[h!]
\centering
\includegraphics[width=0.8\linewidth]{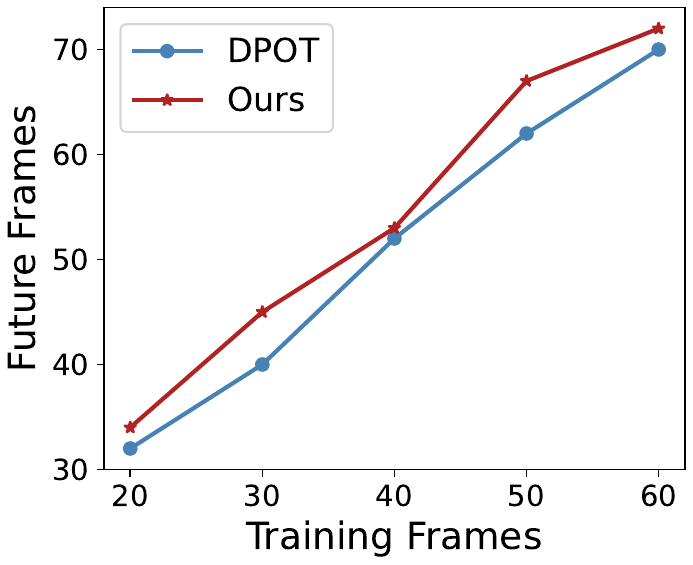}
\centering 
\vspace{-0.5em}
\caption{Comparison of future predictions between our SciML foundation model and DPOT-7M~\cite{hao2024dpot}.
X-axis: different numbers of initial training frames per video.
Y-axis: temporal index of reliably predicted future frames (thresholded by PSNR=25) (higher is better).
}
\label{fig:dpot_future_prediction}
\vspace{-0.5em}
\end{figure}

We further compare with the public available DPOT~\cite{hao2024dpot}, which is a state-of-the-art SciML foundation model.
We choose the DPOT-7M version to match the size of our foundation model (6.5M parameters).
As shown in Figure~\ref{fig:dpot_future_prediction}, future predictions of fine-tuned DPOT-7M
is much worse than our SciML foundation model.

\subsection{Other Baseline}
\label{sec:our_pinf}
We futher test the performance of our method on PINF~\cite{chu2022physics}. Experiments results are shown in Table \ref{tab:pinf_baseline}. Ours-PINF surpasses PINF on all three tasks, showing that our method consistently brings improvements over other Nerf-based baseline. 

In addition, we experiment with using the predictions from HyFluid (v0) as augmented frames for fine-tuning. However, this strategy yields only marginal gains: the future-prediction PSNR increases from 23.91 to 24.91, and the number of predictable future frames increases by only one. 

\begin{table*}[t!]
\centering
\caption{The PSNR gains of using our method over the PINF baseline.
``\texttt{nf}'': number of input training frames.
For future prediction, we report the LPIPS averaged over 20 future frames (i.e., frames with indices from \texttt{nf}+1 $\rightarrow$ \texttt{nf}+20).}
\resizebox{0.75\textwidth}{!}{
\begin{tabular}{lccccccccc}
\toprule
\multirow{2}{*}{Methods} & \multicolumn{3}{c}{Novel View Synthesis} & \multicolumn{3}{c}{Re-Simulation} & \multicolumn{3}{c}{Future Prediction} \\ \cmidrule{2-4} \cmidrule{5-7} \cmidrule{8-10}
 & \texttt{nf}=20 & \texttt{nf}=40 & \texttt{nf}=60 & \texttt{nf}=20 & \texttt{nf}=40 & \texttt{nf}=60 & \texttt{nf}=20 & \texttt{nf}=40 & \texttt{nf}=60 \\ \midrule
PINF~\cite{chu2022physics} & 33.45  & 31.05  & 30.90  & 24.28  & 24.86  & \textbf{24.08}  & 21.71  & 20.85  & 20.67    \\
Ours-PINF &  \textbf{33.81} & \textbf{32.12} & \textbf{31.68}  &  \textbf{25.44}  & \textbf{ 24.87} &  \textbf{24.08}	 & \textbf{26.29} &   \textbf{24.01} & \textbf{24.84}   \\
\bottomrule
\end{tabular}
}
\label{tab:pinf_baseline}
\end{table*}

\section{Implementation Details of SciML Foundation Model}

\subsection{Architecture}

\begin{figure}[h!]
	\centering
	\includegraphics[width=1.0\linewidth]{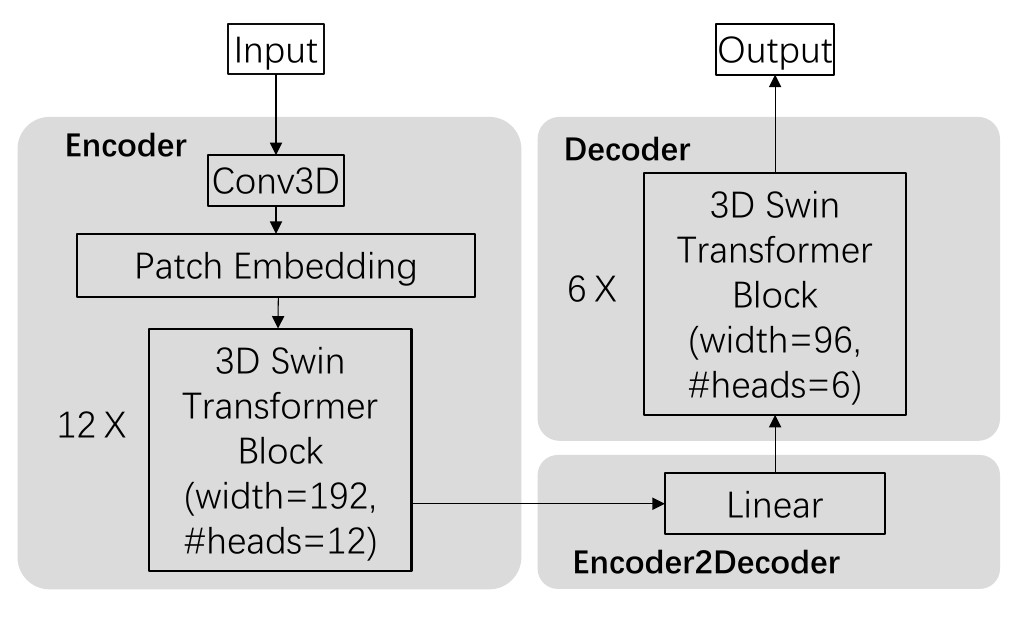}
    \captionsetup{font=small}
\caption{Visualizations of our SciML foundation model architecture (3D Swin Transformer).} 
    \label{fig:swin}
    \vspace{-0.5em}
\end{figure}

To develop our SciML foundation model, we follow the design of 3D Swin Transformer in~\cite{liu2021swin,yang2023swin3d}.
We show our architecture in Figure~\ref{fig:swin}.
Specifically, after the tokenization layer (3D convolution) and patch embedding, we have 18 window attention blocks with a window size of (8,7,7). %

\subsection{Pretraining and Fine-tuning Settings}

We summarize the pretraining and fine-tuning settings of our SciML foundation model in Table~\ref{table:training_settings}.
During both multiphysics pretraining on PDEBench~\cite{PDEBench2022} and fine-tuning on ScalarFlow~\cite{eckert2019scalarflow}, all inputs are interpolated to a spatial resolution of $224\times 224$.
It is worth mentioning that fine-tuning our SciML foundation model only takes 2 hours, much cheaper than 17+ GPU hours of HyFluid (Table \ref{table:psnr_nframes_larger_hyfluid}).

\begin{table*}[h!]
\centering
\caption{Summary of multiphysics pretraining and fine-tuning settings for our SciML foundation model. ``LR'': learning rate. ``BS'': batch size. ``Rollout'': rollout steps for autoregressive predictions. ``v0'' and ``v1'' correspond to fine-tuning foundation models in Figure 5 and 6 in our main paper.}
\resizebox{0.78\textwidth}{!}{
\begin{tabular}{lcccccc}
\toprule
 & LR & BS & Optimizer & Scheduler & Epochs & Rollout \\ \midrule
Pretraining & 0.1 & 4 & SGD & Cosine & 500 & 1 \\
Fine-tuning v0 & 0.1 & 1 & SGD & Cosine & 200 & Increase from 3 to 8 by 1 every 20 epochs \\
Fine-tuning v1 & 0.1 & 1 & SGD & Cosine & 100 & 8 \\ \bottomrule
\end{tabular}
}
\label{table:training_settings}
\end{table*}

\subsection{Feature Extraction for $t < T_{in}$ with Temporal Interpolation}

\begin{figure}[h!]
\centering
\includegraphics[width=1.0\linewidth]{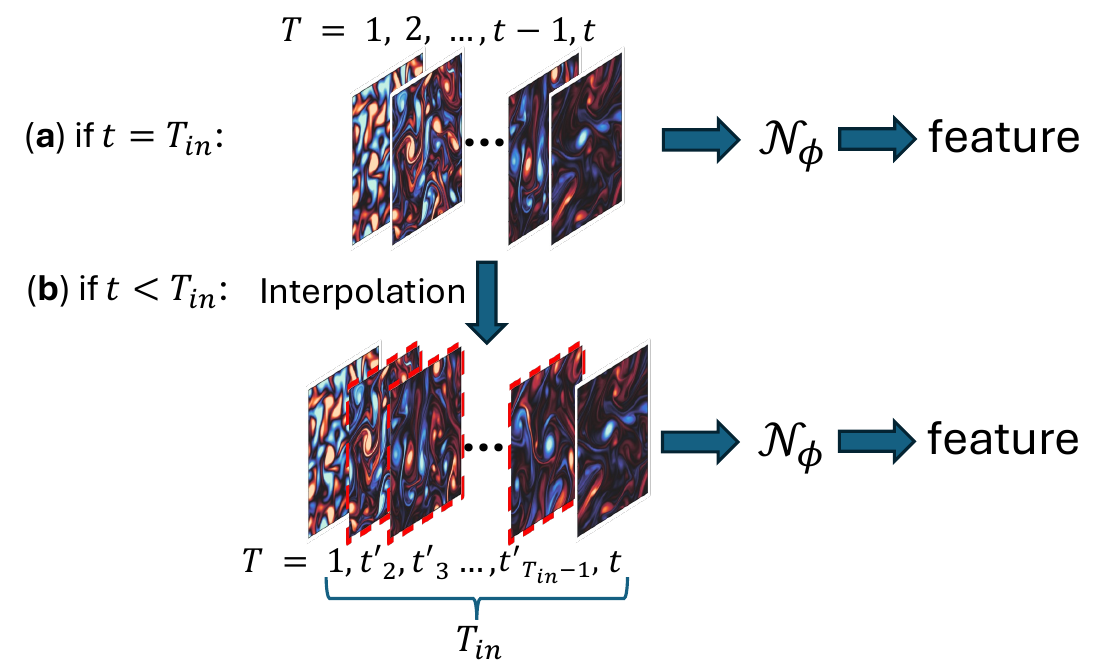}
\centering 
\vspace{-1em}
\caption{Forecasting by SciML foundation models ($\mathcal{N}$ parameterized by $\phi$) ~\cite{mccabe2023multiple,hao2024dpot}. (a) If $t = T_{in}$ (i.e., we have sufficient input frames from previous steps), the model predicts the next step of the fluid dynamics (here, each frame shows the vorticity of the fluid). (b) If $t < T_{in}$ (i.e., we do not have enough input frames requires as inputs), we use temporal-wise interpolation to create $T_{in} - t$ pseudo-frames to supplement necessary frames as inputs to the foundation model.} 
\label{fig:fm_forecasting_interpolation}
\end{figure}

For feature aggregation, during testing, since videos are not accessible, our SciML foundation model extracts features based on frames rendered by the density field from prior temporal steps.
To extract features of frames for temporal steps before $T_{in}$, we use temporal-wise interpolation to supplement necessary frames as inputs to the foundation model.
Specifically, we bi-linearly interpolate the input frames from $t$ to $T_{in}$ steps and create $T_{in} - t$ pseudo-frames, as illustrated in Figure~\ref{fig:fm_forecasting_interpolation}.

\subsection{Pretraining of SciML Foundation Model}
\label{sec:sciml}
\subsubsection{PDEs}
\label{pdes}

We summarize PDEs in our multiphysics pretraining below.

\paragraph{Reaction-Diffusion.}

The Reaction-Diffusion equation models an activator-inhibitor system, which typically happens in the dynamics of chemistry, biology, and ecology.
The activator $u$ promotes its own production,
and the inhibitor $v$ acts to suppress or inhibit the production or activity of the activator.
\begin{align}
 \left\{
    \begin{aligned}
        \partial_t u &= D_u \partial_{x x} u + D_u \partial_{y y} u + R_u,  \\
        \partial_t v &= D_v \partial_{x x} v + D_v \partial_{y y} v + R_v.
    \end{aligned}
    \right.
\end{align}
$D_u,D_v$ are diffusion coefficients ($D_u = 5\times 10^{-3}, D_v = 1\times 10^{-3}$ in PDEBench~\cite{PDEBench2022}). We consider the Fitzhugh-Nagumo type of Reaction-Diffusion equation, which describes how an action potential travels through a nerve membrane ($k$ is set as $5 \times 10^{-3}$):
\begin{equation}
 \left\{
    \begin{aligned} & R_u(u, v)=u-u^3-k-v, \\
    & R_v(u, v)=u-v.
    \end{aligned}
    \right.
\end{equation}

\paragraph{Incompressible Navier-Stokes.}

The Navier–Stokes equations govern the dynamics of fluid flow and serve as fundamentals for fluid simulations. It considers both the mass conservation and the momentum of fluid parcels.
\begin{equation}
    \frac{\partial \mathbf{u}}{\partial t} = - (\mathbf{u} \cdot \nabla) \mathbf{u} + \nu \nabla^2 \mathbf{u}-\frac{1}{\rho} \nabla p,%
    \label{eq:ns}
\end{equation}
where $\mathbf{u}$ is the velocity field, $\nu$ is the dynamic viscosity ($\nu=0.01$ in PDEBench),
$\rho$ is the fluid density (which is constant in incompressible Navier-Stokes),
$p$ is the fluid pressure.
We also have $\nabla \cdot \mathbf{u}=0$ due to the incompressibility.

\paragraph{Compressible Navier-Stokes.}

\begin{table*}[t]
\centering
\captionsetup{font=small}
\caption{Inputs and outputs for learning different PDEs. ``incomp-NS'': incompressible Navier-Stokes. ``CNS'': compressible NS. ``RD'': Reaction-Diffusion. ``SWE'': shallow water equation.
PDE simulations are interpolated to the same spatial resolution $H=W=256$.
}
\resizebox{1.\textwidth}{!}{
\begin{tabular}{lccc}
\toprule
PDE Simulations & Input & Input Shape & Output \\ 
\midrule
RD  & Activator ($u$), inhibitor ($v$) & $T\times C\times H\times W (T=10, C=2)$ & Activator ($u$), inhibitor ($v$) at $T+1$\\ 
incomp-NS  & Velocity ($v_x, v_y$), pressure ($p$) & $T\times C\times H\times W (T=10, C=3)$ & Velocity ($v_x, v_y$), pressure ($p$) at $T+1$ \\ 
CNS  & Velocity ($v_x, v_y$), pressure ($p$), density ($\rho$) & $T\times C\times H\times W (T=10, C=4)$ & Velocity ($v_x, v_y$), pressure ($p$), density ($\rho$) at $T+1$ \\ 
SWE & Height ($h$) & $T\times C\times H\times W (T=10, C=1)$ & Height ($h$) at $T+1$\\ 
Maxwell & Electric field ($E_x$), Magnetic fields ($B_y, B_z$) & $T\times C\times H\times W (T=10, C=3)$ & Electric ($E_x$) and Magnetic fields ($B_y, B_z$) at $T+1$\\ 
\bottomrule
\end{tabular}
}
\label{table:setting_coef_inputoutput}
\end{table*}

\begin{table*}[t]
\centering
\caption{Performance of different SciML foundation models. ``CNS'': compressible Navier-Stokes. ``incomp-NS'': incompressible Navier-Stokes. ``SWE'': shallow water equation. ``RD'': Reaction-Diffusion.}
\resizebox{0.68\textwidth}{!}{
\begin{tabular}{lccccc}
\toprule
Model & \#Parameters (M) & CNS & incomp-NS & SWE & RD \\ \midrule
UNet~\cite{hao2024dpot} & 25 & 0.313 & - & 0.0521 & 0.0971 \\
FNO~\cite{hao2024dpot} & 7 & 0.130 & - & 0.00912 & 0.0321 \\
MPP-Ti~\cite{mccabe2023multiple} & 7 & 0.0442 & - & 0.0066 & 0.0168 \\
DPOT-Ti~\cite{hao2024dpot} & 7 & 0.0285 & - & 0.0056 & 0.0321 \\
Our SciML Foundation Model & 6.5 & 0.195 & 0.094 & 0.0056 & 0.0621 \\ \bottomrule
\end{tabular}
}
\label{table:pde_pretrain}
\end{table*}

The compressible Navier-Stokes equations differ from the incompressible version by accounting for variations in density, pressure, and temperature. The equations include both the conservation of mass (the continuity equation) and conservation of energy, in addition to the momentum conservation equation in the incompressible version.
\begin{align}
    & \text{\hspace{-3.8em} Mass Conservation:}  \notag \\ 
    & \frac{\partial \rho}{\partial t} + \nabla \cdot (\rho \mathbf{u}) = 0, \\
    & \text{\hspace{-3.8em} Momentum Conservation:} \notag \\
    & \frac{\partial (\rho \mathbf{u})}{\partial t} + \nabla \cdot (\rho \mathbf{u} \otimes \mathbf{u}) = -\nabla p + \nabla \cdot \boldsymbol{\tau}, \\
    & \text{\hspace{-3.8em} Energy Conservation:} \notag  \\
    & \frac{\partial (\rho e_t)}{\partial t} + \nabla \cdot \left[ (\rho e_t + p) \mathbf{u} \right] = \nabla \cdot (\mathbf{u} \cdot \boldsymbol{\tau}).
\end{align}

Based on Eq.~\ref{eq:ns}, compressible Navier-Stokes further includes:
$\bm{\tau}$ the viscous stress tensor
$\bm{\tau}=(\zeta+\nu / 3) (\nabla \cdot \mathbf{u})$;
$\nu, \zeta$ shear and bulk viscosity, respectively;
$e_t=e+\frac{1}{2}|\mathbf{u}|^2$ the total energy (internal + kinetic), $e=p /(\Gamma-1)$ and $\Gamma=5 / 3$.
When we define $M=|v| / c_s$ as the Mach number and $c_s=\sqrt{\Gamma p / \rho}$ as the sound velocity,
PDEBench includes two simulation settings: $M=1$ and $M=0.1$.

\paragraph{Shallow Water.}

The shallow water equation (SWE) is a simplified version of the original Navier-Stokes equation, in the context where the scale of horizontal length and velocity is much greater than the vertical scale, and also the pressure distribution is assumed to be hydrostatic.
Typical examples are atmospheric and oceanic modeling.
It is derived by integrating the Navier-Stokes equation over the vertical coordinate from the bottom to the free surface ($h$),
incorporating the gravitational acceleration ($g$),
and replacing horizontal velocities with depth-averaged ones.
\begin{equation}
\begin{gathered}\frac{\partial h}{\partial t}+\nabla \cdot(h \mathbf{\Bar{u}})=0, \quad \frac{\partial \mathbf{\Bar{u}}}{\partial t}+(\mathbf{\Bar{u}} \cdot \nabla) \mathbf{\Bar{u}}=-g \nabla h\end{gathered}
\label{eq:swe}
\end{equation}

In our Shallow Water simulation, our primary interest is in the free surface dynamics ($h$).
This is because, in many practical applications, such as predicting the propagation of surface waves, tsunamis, or tides, the primary quantity of interest is the free surface elevation itself. The horizontal velocity may not be as crucial for these specific phenomena if the primary goal is to understand or predict wave heights, arrival times, or surface oscillations.

In PDEBench, the shallow water simulation considers a 2D radial dam break scenario.
On a square domain $\Omega=[-2.5,2.5]^2$, we initialize the water height as a circular bump in the center of the domain:
\begin{equation}
h(t=0, x, y)= \begin{cases}h_0, & \text { for } \sqrt{x^2+y^2} < r \\ h_1, & \text { for } \sqrt{x^2+y^2} \geq r \end{cases}.
\end{equation}
$h_0 \geq h_1$ and we fix $h_1 = 1$.
In PDEBench, $h_0=2$ and the radius $r$ is sampled from a uniform distribution $\mathcal{U}(0.3,0.7)$.

\paragraph{Maxwell.}

We include the Maxwell's equations not for pretraining our foundation model, but for our ablation study in Section~\ref{sec:exp_benefits_pretraining}.

The Maxwell's equations can calculate the time evolution of an electric field through the following equations:
\begin{equation}
\begin{aligned}
  \nabla \cdot \mathbf{D} &= \rho \\
  \nabla \cdot \mathbf{B} &= 0 \\
  \nabla \times \mathbf{E} &= -\frac{\partial \mathbf{B}}{\partial t} \\
  \nabla \times \mathbf{H} &= \mathbf{J} + \frac{\partial \mathbf{D}}{\partial t}, 
\end{aligned}
\end{equation}
$\mathbf{D}$ is the electric displacement field, $\rho$ is the free charge density, $\mathbf{B}$ is the magnetic flux density, $\mathbf{E}$ is the electric field, $\mathbf{H}$ is the magnetic field, and $\mathbf{J}$ is the free current density.

To simulate Maxwell, we designed a 2D scattering configuration 
to balance simulation speed, storage efficiency, and the complexity of electromagnetic (EM) fields.
The simulation operates in a 2D Y-Z domain, where the X dimension is much smaller and periodic boundary conditions are applied to simulate a 2D scenario effectively.
The simulation area spans $0.002\times 0.002$ meters, featuring a central obstacle with a radius of $2.5\times 10^{-4}$ meters. Four point sources are randomly placed along the edges of the domain, emitting electromagnetic waves. This configuration allows us to capture a rich range of field-matter interactions, enabling both scattering phenomena and field inference dynamics to be observed as they evolve over time.
We use the finite-difference time-domain (FDTD) method to capture the time-dependent behavior of the fields.
We set the relative permittivity $\epsilon_r=50$ for the obstacle.

\paragraph{Summary.}
We summarize details of inputs and outputs of different PDEs in Table~\ref{table:setting_coef_inputoutput}.

\subsubsection{Pretraining Results}

We show results of multiphysics pretraining in Table~\ref{table:pde_pretrain}.
Although results may not be directly comparable due to different training settings, our multiphysics pretraining achieves comparable performance with previous SciML foundation models.